\documentclass[10pt]{article} 
\usepackage[preprint]{tmlr}

\usepackage{amsmath,amsfonts,bm}

\def\eqref#1{equation~\ref{#1}}

\def\1{\bm{1}}

\DeclareMathAlphabet{\mathsfit}{\encodingdefault}{\sfdefault}{m}{sl}
\SetMathAlphabet{\mathsfit}{bold}{\encodingdefault}{\sfdefault}{bx}{n}

\usepackage{hyperref}
\usepackage{url}

\usepackage{float}
\usepackage{amsmath}
\usepackage{graphicx}  
\usepackage{subcaption} 
\usepackage[table, svgnames]{xcolor}
\usepackage{svg}
\usepackage{multirow}
\usepackage{algorithm}
\usepackage{bm}
\usepackage{algpseudocode}
\usepackage{microtype}
\usepackage{hyperref}
\usepackage{url}
\usepackage{booktabs}
\definecolor{darkblue}{rgb}{0, 0, 0.5}
\hypersetup{colorlinks=true, citecolor=darkblue, linkcolor=darkblue, urlcolor=darkblue}
\usepackage{wrapfig}
\usepackage[misc,geometry]{ifsym}
\usepackage{array}
\usepackage{colortbl}

\usepackage{listings}
\usepackage{etoc}
\etocdepthtag.toc{mtchapter}
\etocsettagdepth{mtchapter}{subsection}
\etocsettagdepth{mtappendix}{none}

\title{Beyond Size: How Gradients Shape Pruning Decisions \\ in Large Language Models}

\author{\name Rocktim Jyoti Das\thanks{Equal contribution. $^\dagger$Corresponding author. Now Rocktim is at University of Maryland, College Park, Mukul is at Cornell University, and Mingjie is at Thinking Machines Lab.} \email rocktimjyotidas@gmail.com \\
      \addr Department of Machine Learning \\ Mohamed bin Zayed University of Artificial Intelligence
      \AND
      \name Mukul Ranjan$^*$ \email mukul.ranjan@mbzuai.ac.ae \\
      \addr Department of Machine Learning \\ Mohamed bin Zayed University of Artificial Intelligence
      \AND
      \name Mingjie Sun$^*$ \email mingjies@andrew.cmu.edu\\
      \addr School of Computer Science \\
      Carnegie Mellon University
      \AND
      \name Liqun Ma \email liqun.ma@mbzuai.ac.ae\\
      \addr Department of Machine Learning \\ Mohamed bin Zayed University of Artificial Intelligence
      \AND
      \name Zhiqiang Shen$^{*\dagger}$ \email Zhiqiang.Shen@mbzuai.ac.ae\\
      \addr Department of Machine Learning \\ Mohamed bin Zayed University of Artificial Intelligence}

\def\month{MM}  
\def\year{YYYY} 
\def\openreview{\url{https://openreview.net/forum?id=XXXX}} 

\begin{document}

\maketitle

\begin{abstract}
  Large Language Models (LLMs) with billions of parameters are prime targets for network pruning, which removes some model weights without hurting performance. Prior approaches such as magnitude pruning, SparseGPT, and Wanda either concentrated solely on weights or integrated weights with activations for sparsity. However, they overlooked the informative gradients derived from pretrained LLMs. In this paper, we present a novel sparsity-centric pruning method for pretrained LLMs, termed {\bf G}radient-{\bf b}ased {\bf L}anguage {\bf M}odel {\bf P}runer (\texttt{GBLM-Pruner}). \texttt{GBLM-Pruner} leverages the first-order term of the Taylor expansion, operating in a training-free manner by harnessing properly normalized gradients from a few calibration samples to determine the pruning metric, and substantially outperforms competitive counterparts like SparseGPT and Wanda in multiple benchmarks. Intriguingly, by incorporating gradients, unstructured pruning with our method tends to reveal some structural patterns, which mirrors the geometric interdependence inherent in the LLMs' parameter structure. Additionally, \texttt{GBLM-Pruner} functions without any subsequent retraining or weight updates to maintain its simplicity as other counterparts. {Extensive evaluations on LLaMA-series, Qwen-Series and Deepseek models across zero-shot, STEM-Reasoning and Coding benchmarks show that \texttt{GBLM-Pruner} surpasses magnitude pruning and Wanda by significant margins and performs competitively against SparseGPT which involves intensive weight updates. We further extend our approach on Vision Transformer and Vision Language Models. Our code is available at \url{https://github.com/VILA-Lab/GBLM-Pruner}.}
\end{abstract}

\section{Introduction}
Large Language Models (LLMs) like OpenAI's GPT series~\citep{radford2018improving,radford2019language,brown2020language,openai2023gpt4}, BERT~\citep{devlin2018bert}, LLaMA~\citep{touvron2023llama,touvron2023llama2} and others have made significant strides in recent years, leading to a paradigm shift in natural language processing~\citep{openai2023gpt4,anil2023palm,touvron2023llama2} and multimodal learning~\citep{alayrac2022flamingo,li2023blip}. Many industries have integrated LLMs into their workflows, such as in chatbots~\citep{openai2023gpt4}, 
code completion tools (e.g., GitHub Copilot)~\citep{chen2021evaluating}, and assistive technologies~\citep{zdravkova2022cutting}. While enjoying impressive generalization capabilities, LLMs come with a set of practical challenges and disadvantages. The presence of abundant parameters, large memory consumption, and high computational cost during inference present several concerns in real-world applications. Previous literature proposed multiple solutions to address these disadvantages, such as distillation~\citep{hinton2015distilling}, quantization~\citep{jacob2018quantization}, pruning~\citep{han2015deep}, and hardware acceleration~\citep{chen2020survey}.

Among them, pruning refers to the removal of certain weights or entire neurons/layers based on some criteria, e.g., the smallest weights. A pruned model can maintain similar performance with fewer parameters, resulting in a reduction in storage and computational requirements. Inducing unstructural sparsity in pruning is a widely embraced method aimed at minimizing the memory requirements of neural networks with only a minimal sacrifice in accuracy. Pruning methods stand out as notably simple and efficient mechanisms for model compression, serving to eliminate weights contingent on their significance. Reduced models can be conveniently dispatched to edge devices, and also exhibit substantially lower energy consumption, a sizable portion of energy is expended in transferring model parameters from a device's long-term storage to its memory~\citep{dao2022flashattention}.

However, given the constraints of training-free conditions, existing solutions for pruning LLMs primarily employ either weight magnitude~\citep{han2015learning,han2015deep} or a combination of magnitude and activation ~\citep{Frantar2023SparseGPTML,Sun2023ASA}. While these methods are substantiated with empirical ablations and experiments, they are, to a degree, either too complex to use like SparseGPT by computing matrix inverses and updating weights, or heuristic and lack profound theoretical justification like Wanda, especially concerning the application to the recently developed, highly advanced large language models.

In this study, we tackle the aforementioned complexity and interpretability challenges in LLM pruning methods by presenting a simple yet effective approach named \texttt{GBLM-Pruner} (Gradient-Based Language Model Pruner) that can be theoretically explained using the adapted Optimal Brain Surgeon framework (OBS)~\citep{hassibi1993optimal}. This method proficiently prunes LLMs to significant levels of sparsity, eliminating the need to update the residual weights. 
{\color{black} Specifically, we employ normalization of gradients across calibration samples to formulate an importance indicator matrix $\mathbf{A} \in \mathbb{R}^{d_{\text{out}} \times d_{\text{in}}}$, where $\mathbf{A}[i,j] = \|\mathbf{G}[:,i,j]\|_p$ represents the aggregated gradient magnitude at each weight position. Here, $\mathbf{G}[k,i,j] = \frac{\partial E}{\partial \mathbf{W}[i,j]}\big|_{\text{sample } k}$ is the gradient for calibration sample $k$, and $\|\cdot\|_p$ denotes the $\ell_p$ norm across samples. This gradient-based indicator quantifies weight importance and can either replace or supplement activation-based metrics.
}
This matrix can serve as activations and can either replace or supplement them. This method maintains simplicity over SparseGPT~\citep{Frantar2023SparseGPTML} while showcasing enhanced robustness and improved interpretation than Wanda~\citep{Sun2023ASA} on large language models. Furthermore, it is notable that although we employ gradients in our approach, there is no necessity for retraining or any updates to parameters.

We conducted extensive evaluations of \texttt{GBLM-Pruner} on LLaMA-1 and 2~\citep{touvron2023llama,touvron2023llama2}, among the most influential families of open-source LLMs. Results across various language benchmarks highlight that \texttt{GBLM-Pruner} is proficient in identifying effective sparse networks directly from pretrained LLMs. Notably, \texttt{GBLM-Pruner} substantially surpasses magnitude pruning and recent state-of-the-art methods. Our contributions in this work form a foundational basis for ensuing advancements in this domain.
{\color{black}
Furthermore, we advocate for continued exploration of sparsity within LLMs through the underexplored {\em gradients} angle. While prior gradient-based methods focus on structured pruning or require task-specific finetuning, we introduce the first training-free gradient-based solution for fine-grained unstructured LLM pruning that maintains zero-shot generalization capabilities.} Last, we demonstrate the effectiveness of our approach on Vision Transformers~\citep{dosovitskiy2020image} (see Appendix~\ref{appendix-sec-vit}).

{
Overall, our contributions can be summarized as follows:
\begin{itemize}
\item We present \texttt{GBLM-Pruner}, the first training-free gradient-based 
pruning method for LLMs that maintains task-agnostic zero-shot generalization 
across models up to 70B parameters.
\item We provide theoretical motivation by extending the OBS framework to retain 
first-order gradient information, validated through extensive ablations.
\item We demonstrate consistent improvements over magnitude pruning and Wanda 
(up to 11.19\% relative gain on coding tasks) while being 636$\sim$754$\times$ faster 
than weight update based methods like SparseGPT, with comprehensive evaluation across 15+ model families and 20+ benchmarks.

\end{itemize}
}
\vspace{-1.5ex}
\section{Approach}
{\color{black}
\subsection{Notation}

We establish the following notation used throughout this paper. Matrices are denoted by bold uppercase letters (e.g., $\mathbf{W}$, $\mathbf{X}$), vectors by bold lowercase letters (e.g., $\mathbf{w}$, $\mathbf{g}$), and scalars by regular lowercase letters (e.g., $w_m$, $\alpha$). For a matrix $\mathbf{A}$, we use $\mathbf{A}[i,j]$ to denote the element at row $i$ and column $j$, and $\mathbf{A}[:,j]$ to denote the $j$-th column vector.

\begin{table}[h]
\centering
\small
\color{black} 
\resizebox{0.7\linewidth}{!}{
\begin{tabular}{lll}
\toprule
\textbf{Symbol} & \textbf{Dimension} & \textbf{Description} \\
\midrule
$\mathbf{W}$ & $\mathbb{R}^{d_{\text{out}} \times d_{\text{in}}}$ & Weight matrix of a neural network layer \\
$\mathbf{X}$ & $\mathbb{R}^{N \times L \times d_{\text{in}}}$ & Input activation matrix across samples \\
$\mathbf{G}$ & $\mathbb{R}^{N \times d_{\text{out}} \times d_{\text{in}}}$ & Gradient matrix, $\mathbf{G}[k,i,j] = \frac{\partial E}{\partial \mathbf{W}[i,j]}\big|_k$ \\
$\mathbf{S}$ & $\mathbb{R}^{d_{\text{out}} \times d_{\text{in}}}$ & Importance score matrix (pruning metric) \\
$\mathbf{M}$ & $\{0,1\}^{d_{\text{out}} \times d_{\text{in}}}$ & Binary pruning mask \\
$\mathbf{H}$ & $\mathbb{R}^{d \times d}$ & Hessian matrix, $\mathbf{H} = \frac{\partial^2 E}{\partial \mathbf{w}^2}$ \\
$\mathbf{w}$ & $\mathbb{R}^{d}$ & Vectorized network weights \\
$\mathbf{g}$ & $\mathbb{R}^{d}$ & Gradient vector, $\mathbf{g} = \frac{\partial E}{\partial \mathbf{w}}$ \\
$\delta \mathbf{w}$ & $\mathbb{R}^{d}$ & Change in weight vector \\
$\mathbf{e}_m$ & $\mathbb{R}^{d}$ & Unit vector (1 at position $m$, 0 elsewhere) \\
$N$ & $\mathbb{N}$ & Number of calibration samples \\
$L$ & $\mathbb{N}$ & Sequence length \\
$s$ & $(0,1)$ & Sparsity ratio \\
$\alpha$ & $\mathbb{R}^+$ & Scaling factor for gradient contribution \\
$w_m$ & $\mathbb{R}$ & Scalar weight element at position $m$ \\
$\lambda$ & $\mathbb{R}$ & Lagrange multiplier \\
$E(\cdot)$ & $\mathbb{R}$ & Global loss function \\
$|\cdot|$ & -- & Absolute value (element-wise for matrices) \\
$\|\cdot\|_p$ & -- & $\ell_p$ norm (aggregation across samples) \\
$\|\cdot\|_2$ & -- & $\ell_2$ (Euclidean) norm \\
$(\cdot)^\top$ & -- & Matrix or vector transpose \\
\bottomrule
\end{tabular}
}
\end{table}
}
\vspace{-1.ex}
\subsection{Preliminaries}
\noindent{\bf Weights Magnitude.} Magnitude pruning, which retains weights of significant absolute values, is the predominant approach for weight pruning. It usually induces unstructured sparsity and has been employed across various architectures spanning computer vision~\citep{han2015learning,han2015deep} and language processing~\citep{gale2019state}. Furthermore, it has recently become integral to the lottery ticket hypothesis~\citep{frankle2018lottery}.

\noindent{\bf Weights and Activations.}  SparseGPT~\citep{Frantar2023SparseGPTML} conceptualizes the problem of pruning large language models by addressing a local, layer-wise reconstruction problem. The approach for determining pruning metrics and the process for updating weights in SparseGPT draws inspiration from the Optimal Brain Surgeon (OBS)~\citep{hassibi1993optimal} approach. The pruning metric employed within SparseGPT is defined as follows:

\begin{equation}
{\color{black}\mathbf{S}}[{i, j}]=\frac{|\mathbf{W}[{i, j}]|^2}{\operatorname{diag}\left(\textbf H^{-1}\right)[j,j]}
\end{equation}
where $\textbf H=\left(\mathbf{X}^T \mathbf{X}+\lambda \mathbf{I}\right)$ is the Hessian matrix, and $\textbf H^{-1}$ is the inverse Hessian. ${\color{black}\mathbf{S}}$ is the  importance score for a given weight $\mathbf{W}$, and $[i, j]$ is the element at index $i, j$ of the matrix.

\begin{figure*}[t]
\centering
\includegraphics[width=0.8\linewidth]{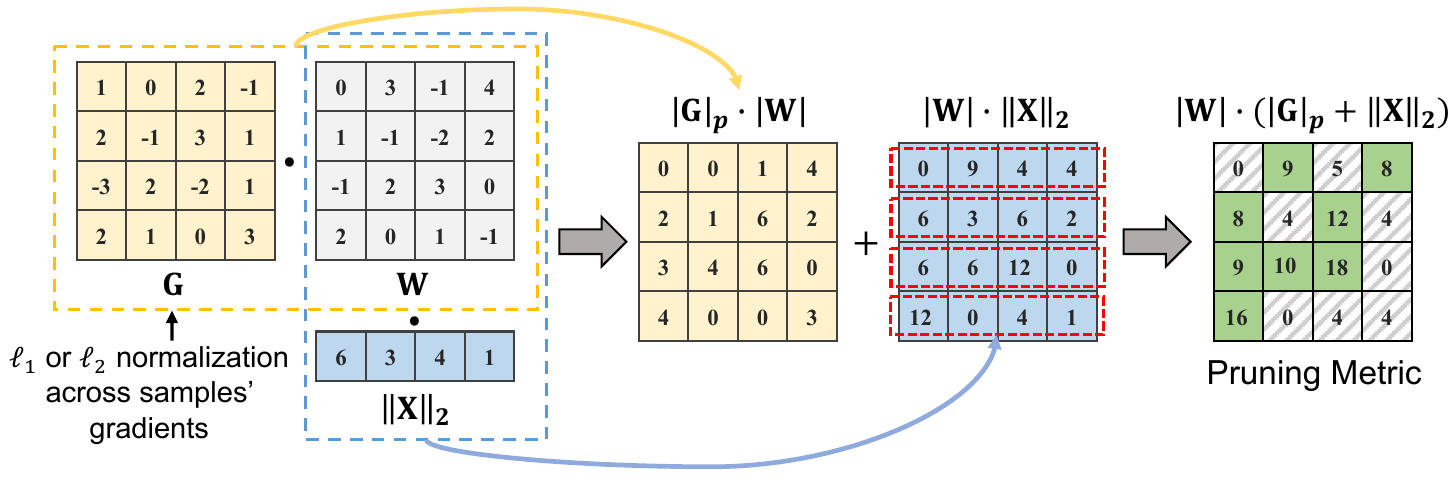}
\vspace{-2.ex}
\caption{Illustration of our method \texttt{GBLM-Pruner}. Given a weight matrix, $\mathbf W$, a gradient matrix, $\mathbf G$, and an input feature activation, $\mathbf X$, weight importance is computed as an element-wise multiplication of weight magnitude and $\ell_1$ or $\ell_2$ norm of the gradients across multiple samples, denoted as \(\|\mathbf G\|_p \cdot  |\mathbf W| \), optionally, it is promotable to add the multiplication of weight and the $\ell_2$ norm of input activations, denoted as \( |\mathbf W| \cdot \|\mathbf X\|_2 \).}
\label{fig:GBLM-Pruner}
\end{figure*}

Wanda~\citep{Sun2023ASA} suggests assessing the significance of each individual weight by calculating the product of its magnitude and the norm of the corresponding input feature. More precisely, the score {$\color{black}\mathbf{S}$} for a given weight, $\mathbf W[i, j]$, is determined as follows:
\begin{equation}
{\color{black}\mathbf{S}}[{i, j}]=\left|\mathbf{W}[{i, j}]\right| \cdot\left\|\mathbf{X}[:,j]\right\|_2
\end{equation}
where the element-wise product between the weight magnitude and the norm of input activations is performed within each row in $\textbf W$.

\begin{algorithm}[!htbp]
\caption{The \texttt{GBLM-Pruner} algorithm}\label{alg:pruning}
\begin{algorithmic}
\State $\textbf W \gets $ weight matrix $\in \left(\bm d_\text{\em out}, \bm d_\text{\em in}\right)$
\State $\textbf X \gets $ activation matrix $\in \left(\bm N\times \bm L, \bm d_\text{\em in}\right)$
\State $\textbf G \gets $ gradient matrix $\in \left(\bm N, \bm d_\text{\em out}, \bm d_\text{\em in}\right)$
{\State $\alpha \gets $ scaling factor for gradient term $\in \mathbb{R}^+$}
\State $\bm s \gets $ sparsity ratio $\in \left(0, 1\right)$
\State ${\color{black}\mathbf{S}} \gets $ pruning metric $\in  \left(\bm d_\text{\em out}, \bm d_\text{\em in}\right)$
\State $\textbf M \gets $ pruning mask $\in  \left(\bm d_\text{\em out}, \bm d_\text{\em in}\right)$\\
\For{ i  $\in \left(1, \bm d_\text{\em out}\right) $}
\For{ j  $\in \left(1, \bm d_\text{\em in}\right) $}
\State{${\color{black}\mathbf{S}}[i,j] = \alpha \cdot \left|\textbf W[i,j]\right|\!\cdot\!\left\|\textbf G[:, i,j]\right\|_{p} \!+\! \left|\textbf W[i,j]\right|\!\cdot\!\left\| \textbf X[:,j]\right\|_2$}
\EndFor
\EndFor\\

\For{ i  $\in \left(1, \bm d_\text{\em out}\right) $}
\State{$\textbf M[i,:]$ = mask of $\bm s\%$ weights with smallest ${\color{black}\mathbf{S}}[i,:]$}
\EndFor\\
\State $\textbf W[\textbf M] = 0$
\end{algorithmic}
\end{algorithm}
\subsection{Gradients Matter}
\noindent{\bf Gradients.} {According to Optimal Brain Damage~\citep{lecun1989optimal} and Optimal Brain Surgeon~\citep {hassibi1993optimal}, gradients and higher order derivatives are naturally correlated to the importance of weights for LLM pruning, which is the theoretical basis of our approach. However, they ignore the gradients in their pruning framework under the assumption that gradients of the fully trained network are small and do not provide any additional information when the higher-order terms are considered. Our work shows that gradients are still crucial and provide non-trivial information.

Previous gradient-based structured pruning methods, such as feature map pruning~\citep{molchanov2016pruning},
channel pruning~\citep{yang2022gradient}, and head pruning~\citep{michel2019sixteen} utilize the first-order Taylor approximation of loss $\mathcal{L}$ around activation $z=0$ or weight $w=0$ as importance score, the formulation is:
\begin{equation}
{\color{black}\mathbf{S}}=\mathbb{E}_{\bm x \sim \textbf X}\left|\frac{\partial \mathcal{L}(\bm x)}{\partial \textbf A} \textbf A\right|
\end{equation}
where $\textbf X$ is the sampled data distribution and $A$ is either activation matrix $Z$ or weight matrix $W$. Most of these structured pruning methods are proposed for transfer learning to a particular task and require significant finetuning on the specific task to maintain the model performance. In contrast, our work leverages gradient information to do unstructured and N$:$M semi-structured pruning without any subsequent weight update. Additionally, we illustrate the integration of activations into our pruning metric through the use of a scaling factor for best performance. Furthermore, our pruned model is task-agnostic and generalizable to any downstream task as showcased by the Zero-shot evaluation on several tasks included in the {\color{black}Eleuther} AI lm-evaluation harness benchmark~\citep{eval-harness}.

\noindent{\bf Pruning Metric.} Consider a layer in LLMs characterized by the weight \( \textbf W \), possessing a shape of \( (\bm d_\text{\em out}, \bm d_\text{\em in}) \). In the context of Transformer models, this layer has the gradient $\textbf G$, exhibiting the same shape of weight \( \textbf W \). We propose evaluating the importance of each individual weight by aggregating the corresponding gradients across different samples and then computing the product of its magnitude with the weights. More precisely, the importance score attributed to a specific weight, \(\textbf W[i, j] \), is determined as follows:
\begin{equation}
{\color{black}\mathbf{S}}[i,j] = \left|\textbf W[i,j]\right|\cdot\left\| \textbf  G[:, i,j]\right\|_{p}
\label{eq:grad_weight}
\end{equation}
While competitive results can be achieved with gradients solely, we can combine feature activations to get better performance, which forms our final pruning metric:
\begin{align}
    {\color{black}\mathbf{S}}[i,j] =  \alpha \cdot \left|\textbf W[i,j]\right| \cdot \left\|\textbf G[:, i,j]\right\|_{p}  +  \left|\textbf W[i,j]\right|\cdot \left\| \textbf X[:,j]\right\|_2
\label{eq:final_prun_metric}
\end{align}

where $\alpha$ is the scaling factor used to account for the small magnitude of gradients, which makes the contribution of gradient balanced to the large magnitude of activations. The $\alpha$ is chosen using a held-out validation set.

\noindent{\bf Comparison Group.} The comparison group of pruning is pivotal in unstructured pruning, owing to the fact that varying granularities yield disparate pruning patterns. Previously, unstructured magnitude pruning approaches have leveraged both layer-wise and global pruning. In these methods, weights are contrasted either within the same layer or throughout the entirety of the model. Through a comprehensive study, we observe that the highest accuracy is achieved when weights are analyzed on a {\color{black}row-wise} basis. 
This insight is consistent with the findings  in~\cite{Sun2023ASA}.

We illustrate in detail our proposed pruning approach \texttt{GBLM-Pruner} in Algorithm~\ref{alg:pruning}.

{\color{black}\subsection{A Theoretical Motivation}\label{sec:theory}}

In this section, we have revisited and refined the Optimal Brain Surgeon (OBS) framework~\citep{hassibi1993optimal} framework by incorporating considerations of the gradient, i.e., the first-order term in Taylor approximation. The closed-form solution of the increase in error for removing a weight from the model, given by this analysis serves as the fundamental basis for our novel gradient-based pruning metric. For the sake of simplicity, we will consider weights and gradients as one-dimensional vectors denoted by $\bm w$ and $\bm g$ respectively in our analysis.

{\color{black}
We work with the squared error loss function:
\begin{equation}
E = \frac{1}{2N}\sum_{k=1}^{N} (\mathbf{t}^{[k]} - \mathbf{o}^{[k]})^{\top}(\mathbf{t}^{[k]} - \mathbf{o}^{[k]})
\label{eq:loss_function}
\end{equation}
where $N$ is the number of calibration samples, $\mathbf{t}^{[k]}$ is the target output, and $\mathbf{o}^{[k]}$ is the network output for sample $k$.
}
The optimization problem for network pruning using both the first and second-order terms can be depicted in Equation \ref{eq:prun_opt_obj}. Here, $\bm E$ is the error or loss function, $\bm w$ is the weight vector for the neural network, and $\delta \bm w$ is the change in the weight vector. Additionally, $ \mathbf{e}_{m}$ is the unit vector in weight space corresponding to the pruned weight $w_m$, $ \textbf H=\frac{\partial^2 \bm E}{\partial \bm w^2}$ denotes the Hessian Matrix, and the superscript $\top$ signifies vector transpose.

\begin{equation}
\begin{aligned}
    \min_{m}\Bigg\{ &\min_{\delta \bm w}\left(\left(\frac{\partial \bm{E}}{\partial \bm w}\right)^{\top}\cdot\delta \bm{w} \right. \\
    &\quad \left. + \frac{1}{2} \delta \bm{w}^{\top} \cdot \textbf{H} \cdot \delta \bm{w} \right) \Big| \mathbf{e}_m^{\top} \cdot \delta \bm{w} + w_m = 0\Bigg\}
\end{aligned}  
\label{eq:prun_opt_obj}
\end{equation}

{\color{black}
The constraint $\mathbf{e}_{m}^{\top} \cdot \delta \mathbf{w} + w_m = 0$ enforces that after weight adjustment, the pruned weight $w_m$ is set to exactly zero. Specifically, if the new weights are $\mathbf{w}' = \mathbf{w} + \delta \mathbf{w}$, we require $w'_m = 0$, which gives $\delta w_m = -w_m$. Since $\mathbf{e}_{m}$ is a unit vector with 1 at position $m$ and 0 elsewhere, we have $\mathbf{e}_{m}^{\top} \delta \mathbf{w} = \delta w_m$, thus $\mathbf{e}_{m}^{\top} \delta \mathbf{w} + w_m = 0$.

By solving the optimization problem, we obtain the optimal change in error, $\delta \bm E_m$, for removing weight $w_m$ as shown in Equation \ref{eq:lag_sol}. We provide detailed derivation in Appendix \ref{app:obs}.
}

\begin{equation}
\begin{aligned} 
\delta \bm E_m &= \frac{w_{m}^{2}}{2\left(\textbf{H}^{-1}\right)_{mm}} - \frac{w_{m}\left(\bm{g}^{\top} \cdot \textbf{H}^{-1} \cdot \mathbf{e}_{m}\right)}{\left(\textbf{H}^{-1}\right)_{mm}} \\
&\quad + \frac{\left(\mathbf{e}_{m}^{\top} \cdot \textbf{H}^{-1} \cdot \bm{g}\right)^{2}}{2\left(\textbf{H}^{-1}\right)_{mm}} - \frac{1}{2} \bm{g}^{\top} \cdot \textbf{H}^{-1} \cdot \bm{g}
\end{aligned}
\label{eq:lag_sol}
\end{equation}

For the error, $\delta \bm E_m$, since the gradients are already small, we can consider the quadratic or square term of the gradient to be insignificant. Thus, ignoring the third and fourth terms, we have:
\begin{equation}
\begin{aligned}
\delta \bm E_m & =\frac{w_{m}^{2}}{2\left( \textbf H^{-1}\right)_{m m}}-\frac{{w}_{m}\left(\bm{g}^{\top} \cdot \textbf{H}^{-1} \cdot \mathbf{e}_{m}\right)}{\left({\textbf H}^{-1}\right)_{m m}}  
\end{aligned}
\label{eq:lag_sol_approx}
\end{equation}

{\color{black}
\textbf{Hessian Approximation.} To compute the Hessian matrix $\mathbf{H}$ of the global loss function, we follow the approximation from the original OBS framework~\citep{hassibi1993optimal}. Under the $(t-o) \to 0$ approximation for well-trained networks, OBS derives (their Equation 11):
\begin{equation}
\mathbf{H} \approx \frac{1}{N}\sum_{k=1}^{N} \mathbf{J}^{[k]\top}\mathbf{J}^{[k]}
\label{eq:hessian_obs}
\end{equation}
where $\mathbf{J}^{[k]} = \frac{\partial \mathbf{F}(\mathbf{w}, \mathbf{x}^{[k]})}{\partial \mathbf{w}}$ is the Jacobian of network outputs with respect to weights for sample $k$.

For layer-wise analysis of a linear transformation $\mathbf{a}^{[k]} = \mathbf{W}\mathbf{x}^{[k]}$, the Jacobian with respect to weight $W[i,j]$ involves the input activation $x_j^{[k]}$. Through the chain rule, this yields $\mathbf{H} \approx \mathbf{X}^{\top}\mathbf{X}$ where $\mathbf{X}$ contains the input activation features across samples. We provide the complete derivation in Appendix~\ref{app:hessian_derivation}.

We make two simplifying assumptions to enable tractable computation:

\textbf{Assumption A1 (Diagonal across neurons):} Following Optimal Brain Damage~\citep{lecun1989optimal}, we assume interactions between different output neurons in the same layer are negligible, giving a block-diagonal Hessian structure.

\textbf{Assumption A2 (Uniform curvature):} We approximate the second-order derivatives $\frac{\partial^2 E}{\partial a_i^2}$ as approximately constant across neurons within a layer. While this is a strong assumption that may not hold exactly, it enables computational efficiency. The same approximations are implicitly made in prior accepted work such as Wanda~\citep{Sun2023ASA}, whose metric omits curvature terms entirely.

Under these assumptions, we restrict our focus to the diagonal elements:
\begin{equation}
\textbf{H} \approx 2 \cdot \operatorname{diag}\left(\left\{\left\|\bm{x}_{j}\right\|_{2}^{2}, 1 \leq j \leq n\right\}\right)
\label{eq:hessian_diagonal}
\end{equation}
where $\bm{x}_j$ is the activation vector corresponding to component $j$ across samples, and $n$ is the total number of components within the activation tensor for the layer.

The first term of Equation \ref{eq:lag_sol_approx} then becomes:
}

\begin{equation}
\begin{aligned}
\frac{{w}_{m}^2}{2\left(\textbf{H}^{-1}\right)_{m m}} & ={w}_{m}^2\left\| \bm{x}_m\right\|_2^{2}
\end{aligned}
\label{eq:lag_soln_first_term}
\end{equation}

{\color{black}
Similarly, considering only diagonal elements and using the diagonal Hessian approximation, the second term in Equation \ref{eq:lag_sol_approx} simplifies to:
}

\begin{equation}
\begin{aligned}
-\frac{{w}_{m}\left(\bm{g}^{\top} \cdot \textbf{H}^{-1} \cdot \mathbf{e}_{m}\right)}{\left(\textbf{H}^{-1}\right)_{m m}}= -\frac{{w}_{m} g_m\left(\textbf{H}^{-1}\right)_{m m}}{\left(\textbf{H}^{-1}\right)_{m m}}={w}_{m}(-{g}_{m})
\end{aligned}
\label{eq:lag_soln_second_term}
\end{equation}

Thus, the final solution for the optimization problem in Equation \ref{eq:prun_opt_obj} can be expressed as:
\begin{equation}
\begin{aligned}
\delta \bm E_m=\left({w}_{m}\left\| \bm{x}_m\right\|_2\right)^{2}+{w}_m\left(-{g}_m\right)
\end{aligned}
\label{eq:final_soln}
\end{equation}

{\color{black}
Equation \ref{eq:final_soln} provides theoretical motivation for combining weight magnitude, activation norms, and gradient information. However, through extensive empirical investigation in Section~\ref{sec:abl}, we find that the following practical formulation yields superior performance:
\begin{equation}
\mathbf{S}[i,j] = |\mathbf{W}[i,j]| \cdot \|\mathbf{X}[:,j]\|_2 + \alpha \cdot |\mathbf{W}[i,j]| \cdot \|\mathbf{G}[:,i,j]\|_p
\label{eq:practical_metric}
\end{equation}
where $\mathbf{S}$ is the importance score matrix, $\alpha$ is an empirically determined scaling factor (we use $\alpha=100$ based on validation experiments in Table~\ref{tab:scale_factor}), and $\|\mathbf{G}[:,i,j]\|_p$ represents the $\ell_p$ norm of gradients across samples.

The key modifications from the theoretical form are: (1) removing the square from the activation term, (2) introducing absolute values for stability, and (3) using $\ell_p$ norm aggregation of gradients rather than individual gradient values. These adjustments improve empirical performance while preserving the core theoretical insight: weight importance should balance gradient magnitude with parameter sensitivity to input activations.
}

\section{Experiments}

\begin{table*}[t]
\centering
\caption{Zero-Shot evaluation on 50$\%$ unstructured sparsity pruned models for different benchmarks.}
\label{tab:harness_combined}
\resizebox{0.88\textwidth}{!}{
\begin{tabular}{clcccccc|c}
\toprule
\multicolumn{1}{l}{Models} & Method & \multicolumn{1}{c}{BoolQ} & \multicolumn{1}{c}{RTE} & \multicolumn{1}{c}{HellaSwag} & \multicolumn{1}{c}{WinoGrande} & \multicolumn{1}{c}{ARC-e} & \multicolumn{1}{c}{OBQA} & \multicolumn{1}{|c}{Mean} \\ \hline
\multirow{5}{*}{LLaMA-1-7B} & Dense & 75.11 & 66.43 & 76.21 & 69.85 & 72.81 & 44.40 & 67.47 \\  
 & Magnitude & 54.65 & 54.15 & 60.90 & 59.43 & 54.38 & 35.20 & 53.12 \\  
 & SparseGPT & \underline{72.87} & 53.07 & 69.77 & \textbf{67.88} & \textbf{66.46} & \underline{40.60} & \underline{61.77} \\  
 & Wanda & 71.25 & \underline{54.87} & \underline{70.12} & 66.06 & 65.11 & 39.60 & 61.17 \\ 
  \rowcolor{Gainsboro!40} & Ours & \textbf{73.43} & \textbf{59.93} & \textbf{70.29} & \underline{67.40} & \underline{65.99} &\textbf{41.40} & \textbf{63.07} \\ \hline
\multirow{5}{*}{LLaMA-1-13B} & Dense & 77.98 & 70.40 & 79.07 & 72.77 & 74.75 & 44.80 & 69.96 \\ 
 & Magnitude & 54.95 & 50.90 & 59.69 & 63.54 & 54.25 & 39.80 & 53.86 \\ 
 & SparseGPT & \textbf{76.67} & 63.18 & 74.09 & 71.59 & 68.48 & \textbf{43.60} & 66.27 \\  
 & Wanda & 76.02 & 63.18 & \underline{74.80} & \textbf{71.90} & \underline{69.82} & 43.00 & \underline{66.45} \\ 
 \rowcolor{Gainsboro!40} & Ours & \underline{76.61} & \textbf{63.18} & \textbf{74.90} & \underline{71.67} & \textbf{70.37} & \underline{43.20} & \textbf{66.65} \\ \hline
\multirow{5}{*}{LLaMA-1-30B} & Dense & 82.72 & 66.79 & 82.62 & 75.77 & 78.91 & 48.20 & 72.50 \\ 
 & Magnitude & 64.25 & 49.82 & 67.29 & 66.61 & 70.71 & 41.20 & 59.98 \\  
 & SparseGPT & \textbf{82.91} & 55.96 & 79.31 & \textbf{74.27} & \underline{77.53} & 46.00 & 69.33 \\ 
 & Wanda & 81.71 & \underline{65.34} & \underline{79.91} & 73.56 & \textbf{78.11} & \textbf{46.40} & \underline{70.84} \\  
 \rowcolor{Gainsboro!40} & Ours & \underline{82.69} & \textbf{67.15} & \textbf{80.23} & \underline{73.95} & 76.98 & \underline{46.00} & \textbf{71.17} \\ \hline
\multirow{5}{*}{LLaMA-2-13B} & Dense & 80.55 & 65.34 & 79.38 & 72.22 & 77.44 & 45.20 & 70.02 \\ 
 & Magnitude & 57.65 & 55.96 & 73.02 & 65.35 & 67.17 & 40.80 & 59.99 \\ 
 & SparseGPT & \textbf{81.25} & \textbf{62.82} & 75.34 & 70.48 & 71.34 & 44.00 & 67.54 \\ 
 & Wanda & \underline{81.07} & 60.65 & \textbf{76.08} & \underline{71.67} & \underline{71.63} & \underline{44.60} & \underline{67.62} \\ 
 \rowcolor{Gainsboro!40} & Ours & 80.89 & \underline{60.65} & \underline{76.03} & \textbf{71.82} & \textbf{72.26} & \textbf{44.80} & \textbf{67.74} \\ \hline 
\multirow{4}{*}{Qwen3-14B} & Dense & 89.33 & 77.62 & 78.86 & 73.09 & 82.83 & 46.20 & 74.65 \\
 & SparseGPT & 87.86 & 79.06 & \textbf{75.24} & 71.90 & 74.75 & \textbf{45.60} & 72.40 \\
 & Wanda & \textbf{88.04} & \textbf{80.87} & 73.00 & \underline{72.45} & \textbf{79.80} & 43.00 & \underline{72.86} \\
 \rowcolor{Gainsboro!40} & Ours & \underline{87.95} & \underline{80.51} & \underline{74.86} & \textbf{72.61} & \underline{79.25} & \underline{43.80} & \textbf{73.16} \\ 
\bottomrule
\end{tabular}
}
\vspace{-0.1in}
\end{table*}

\subsection{Implementation and Setup Details} \label{exp_details} 
We conduct all our experiments using PyTorch \citep{Paszke2017AutomaticDI} for \texttt{GBLM-Pruner}.
 Experiments are performed with five models from the LLaMA-1 series (7B, 13B, 30B)~\citep{touvron2023llama}, the LLaMA-2-13B~ \citep{touvron2023llama2} and the Qwen3-14B~ \citep{yang2025qwen3}. The Huggingface transformer library is used \citep{Wolf2019HuggingFacesTS} for handling models. The experiments are conducted on NVIDIA A100 GPUs with 40/80GB of memory. \texttt{GBLM-Pruner} requires calibration data for the computation of gradients and activations. Following previous works \citep{Frantar2022GPTQAP, Frantar2023SparseGPTML, Sun2023ASA}, we use 128 sequences with 2048 tokens randomly sampled from the first shard of the C4~\citep{Raffel2019ExploringTL} training data as our calibration data. The gradients are computed with language modeling on the input sequence as the objective function. This represents the pretraining objective of the language models and remains agnostic to the downstream tasks for which the language models are used. For scaling factor $\alpha$, we use 100 after ablation using a validation set.

\noindent{\bf Evaluation.} 
We assess the performance of the pruned models using two distinct metrics: {\bf (1)} Perplexity and {\bf (2)} Zero-shot Evaluation on the Harness Benchmark~\citep{eval-harness}. Perplexity is a well-established metric~\citep{Dettmers2022TheCF, Yao2022ZeroQuantEA, Frantar2022OptimalBC, Sun2023ASA, Frantar2023SparseGPTML} and provides stable and reliable results. The Zero-shot Harness evaluation, although known to be relatively noisy, offers a more readily interpretable assessment of model performance.

\begin{table}[!htbp]
\centering
\vspace{-0.08in}
\caption{Comparison group for \texttt{GBLM-Pruner}.}
\vspace{1.ex}
\label{tab:prune_config}
\small
\begin{tabular}{lc}
\toprule
Comparison Group & \multicolumn{1}{c}{Perplexity} \\ \midrule 
\texttt{layer} & \underline{7.45}  \\ 
\texttt{input,1} & 10.16\ \; \\ 
\texttt{input,128} & 7.64  \\ 
\texttt{output,1} & \bf 6.86  \\ 
\texttt{output,128} & 7.47  \\
\bottomrule
\end{tabular}
\vspace{-0.1in}
\end{table}

\noindent{\bf Sparsity and Comparison Group.}
Following recent methods~\citep{Frantar2023SparseGPTML, sanh2020movement}, \texttt{GBLM-Pruner} prunes the linear layers of LLMs uniformly except for the embedding layer and the final classification head. In addition to unstructured pruning, we also position \texttt{GBLM-Pruner} in comparison to other baselines, exploring more rigorous yet hardware-accommodating 2:4 and 4:8 semi-structured sparsity patterns. 
We experiment with five different pruning configurations, as shown in Table \ref{tab:prune_config}. Our findings indicate that the (\texttt{output,1}) configuration yields the most favorable results, prompting its adoption as the standard for all our experiments.

\vspace{-0.06in}
\subsection{Perplexity Evaluation}
\vspace{-0.06in}
For all the comparing methods, we report the perplexity evaluated on WikiText~\citep{Merity2016PointerSM} validation data for both unstructured and semi-structured N$:$M sparsity pruning in Table~\ref{tab:ppl_result_llama1_llama2}. For unstructured pruning, \texttt{GBLM-Pruner} with $\ell_1$ norm outperforms both Wanda and reconstruction-based SparseGPT significantly across both LLaMA-1 and LLaMA-2 models. However, the N$:$M sparsity pruning is restrictive by definition, especially 2:4 sparsity, which imposes greater constraints and results in a noticeable increase in perplexity compared to unstructured pruning. As shown in Table \ref{tab:ppl_result_llama1_llama2}, we can observe SparseGPT seems to perform better than both \texttt{GBLM-Pruner} and Wanda in the case of 2:4 sparsity pruning.  Conversely, for 4:8 sparsity pruning, \texttt{GBLM-Pruner} outperforms other baselines for most of models, especially for the larger models. 

\begin{table}[!htbp]
\caption{WikiText perplexity of pruned LLaMA-1 and LLaMA-2 family of models.}
\label{tab:ppl_result_llama1_llama2}
\vspace{-0.05in}
\renewcommand{\arraystretch}{1.15}
\begin{center}
\resizebox{0.65\textwidth}{!}{
\begin{tabular}{lcccc|ccc}
\toprule
 &  & \multicolumn{3}{c|}{LLaMA-2}  & \multicolumn{3}{c}{LLaMA-1} \\ \hline
Method & Sparsity & \multicolumn{1}{c}{7B} & \multicolumn{1}{c}{13B} & 70B & \multicolumn{1}{c}{7B} & \multicolumn{1}{c}{13B} & \multicolumn{1}{c}{30B} \\ \hline
None & \multicolumn{1}{c}{0} & 5.47 & 4.88 & 3.32  & \multicolumn{1}{r}{5.68} & \multicolumn{1}{r}{5.09} & \multicolumn{1}{r}{4.10} \\ \hline
Magnitude & \multicolumn{1}{c}{0.5} & 16.03 & 6.83 & 5.36 & \multicolumn{1}{r}{17.29} & \multicolumn{1}{r}{20.21} & \multicolumn{1}{r}{7.54}  \\ 
SparseGPT & \multicolumn{1}{c}{0.5} & 7.00 & 6.03 & 4.25 & \multicolumn{1}{r}{7.22} & \multicolumn{1}{r}{6.19} & \multicolumn{1}{r}{5.32}  \\ 
Wanda & \multicolumn{1}{c}{0.5} & \underline{6.92} & \underline{5.97} & \underline{4.22} & \multicolumn{1}{r}{\underline{7.26}} & \multicolumn{1}{r}{\underline{6.15}} & \multicolumn{1}{r}{\underline{5.24}} \\ 
\rowcolor{Gainsboro!40}\texttt{GBLM-Pruner}$_{\ell1}$ & \multicolumn{1}{c}{0.5} & \textbf{6.86} & \textbf{5.88} & \textbf{4.17} & \multicolumn{1}{r}{\textbf{7.15}} & \multicolumn{1}{r}{\textbf{6.11}} & \multicolumn{1}{r}{\textbf{5.18}}  \\ \hline
Magnitude & 2:4 & 37.77 & 8.89 & 6.76  & \multicolumn{1}{r}{42.54} & \multicolumn{1}{r}{18.36} & \multicolumn{1}{r}{9.11} \\ 
SparseGPT & 2:4 & \textbf{10.82} & \textbf{8.75} & 5.68 & \multicolumn{1}{r}{\textbf{10.88}} & \multicolumn{1}{r}{\textbf{9.06}} & \multicolumn{1}{r}{7.12}\\ 
Wanda & 2:4 & 12.11 & 9.00 & \underline{5.48} & \multicolumn{1}{r}{11.53} & \multicolumn{1}{r}{9.59} & \multicolumn{1}{r}{\underline{6.90}}  \\ 
\rowcolor{Gainsboro!40}\texttt{GBLM-Pruner}$_{\ell1}$  & 2:4 & \underline{11.91} & \underline{8.80} & \textbf{5.47} & \multicolumn{1}{r}{\underline{11.33}} & \multicolumn{1}{r}{\underline{9.16}} & \multicolumn{1}{r}{\textbf{6.87}}   \\ \hline
Magnitude & 4:8 & 15.91 & 7.32 & 5.89  & \multicolumn{1}{r}{16.83} & \multicolumn{1}{r}{13.87} & \multicolumn{1}{r}{7.62}  \\ 
SparseGPT & 4:8 & \textbf{8.46} & 7.01 & 4.91 & \multicolumn{1}{r}{\textbf{8.45}} & \multicolumn{1}{r}{7.44} & \multicolumn{1}{r}{6.18} \\ 
Wanda & 4:8 & \underline{8.60} & \underline{7.00} & \underline{4.77} & \multicolumn{1}{r}{8.57} & \multicolumn{1}{r}{\underline{7.41}} & \multicolumn{1}{r}{\underline{5.97}}   \\ 
\rowcolor{Gainsboro!40}\texttt{GBLM-Pruner}$_{\ell1}$ & 4:8 & 8.63 & \textbf{6.90} & \textbf{4.72} & \multicolumn{1}{r}{\underline{8.48}} & \multicolumn{1}{r}{\textbf{7.26}} & \multicolumn{1}{r}{\textbf{5.89}}  \\ 
\bottomrule
\end{tabular}
}
\end{center}
\vspace{-0.1in}
\end{table}

\begin{table}[t]
\centering
\caption{Performance comparison of base and pruned language models with 50\% unstructured sparsity on STEM and general purpose reasoning benchmarks. }
\label{tab:math_science_performance}
\resizebox{0.75\textwidth}{!}{%
\begin{tabular}{@{}ll*{7}{c}@{}}
\toprule
\textbf{Model} & \textbf{Methods} & \rotatebox{0}{\textbf{MMLU }} & \rotatebox{0}{\textbf{GPQA-D}} & \rotatebox{0}{\textbf{GPQA}} & \rotatebox{0}{\textbf{MATH}} & \rotatebox{0}{\textbf{GSM8K}} & \rotatebox{0}{\textbf{Mean}} \\
   &     &   (5-s)   & (5-s)  &  (5-s) & (4-s) & (5-s) &  \\ 
\midrule
\multirow{4}{*}{Llama-3.1-8B} & Base & 65.20 & 24.24 & 27.68 & 8.54 & 51.55 & 35.44 \\
& SparseGPT & \textbf{53.21} & 23.23 & 22.32 & 7.76 & \textbf{16.60} & \underline{24.63} \\
& Wanda & 49.34 & \underline{24.18} & \underline{23.66} & \underline{7.92} & 13.57 & 23.73 \\
\rowcolor{Gainsboro!40} & Ours & \underline{50.16} & \textbf{25.25} & \textbf{25.00} & \textbf{8.54} & \underline{14.86} & \textbf{24.76} \\
\midrule
\multirow{4}{*}{Qwen3 8B} & Base & 74.95 & 19.70 & 28.57 & 25.70 & 87.34 & 47.25 \\
& SparseGPT & \textbf{69.04} & 21.72 & \textbf{25.89} & 16.74 & 73.09 & 41.30 \\
& Wanda & \underline{67.36} & \textbf{26.77} & 24.33 & \underline{17.02} & \underline{73.62} & \underline{41.82} \\
\rowcolor{Gainsboro!40} & Ours & 66.75 & \underline{26.26} & \underline{25.67} & \textbf{17.18} & \textbf{74.15} & \textbf{42.00} \\
\midrule
\multirow{4}{*}{Qwen3-14B} & Base & 78.81 & 23.23 & 26.79 & 32.50 & 84.84 & 49.23 \\
& SparseGPT & \textbf{74.29} & \textbf{31.82} & 23.88 & 22.78 & 76.35 & 45.82 \\
& Wanda & \underline{74.07} & 25.25 & \textbf{29.69} & \underline{23.54} & \underline{79.00} & \underline{46.31} \\
\rowcolor{Gainsboro!40} & Ours & 74.01 & \underline{29.80} & \underline{29.46} & \textbf{23.80} & \textbf{80.82} & \textbf{47.58} \\
\midrule
\multirow{4}{*}{Qwen2.5 7B} & Base & 74.34 & 24.24 & 25.22 & 24.88 & 82.18 & 46.17 \\
& SparseGPT & \textbf{67.47} & \textbf{26.30} & 23.66 & 15.82 & 58.15 & \underline{38.28} \\
& Wanda & 66.76 & 20.20 & \underline{23.88} & \underline{16.50} & \underline{60.27} & 37.52 \\
\rowcolor{Gainsboro!40} & Ours & \underline{67.16} & \underline{21.72} & \textbf{25.22} & \textbf{16.70} & \textbf{62.24} & \textbf{38.61} \\
\bottomrule
\end{tabular}%
}\vspace{-0.15in}
\end{table}

\begin{table}[t]
\centering
\caption{Performance comparison of base and pruned language models on coding benchmarks. All pruned models use 50\% unstructured sparsity. }
\label{tab:model_performance}
\resizebox{0.65\textwidth}{!}{
\begin{tabular}{@{}llccc@{}}
\toprule
\textbf{Model} & \textbf{Method} & \textbf{MBPP} & \textbf{HumanEval} & \textbf{Mean} \\
& & \textbf{(3-shot)} & \textbf{(0-shot)} & \\
\midrule
\multirow{4}{*}{Qwen3-8B} & Base & 65.60 & 62.80 & 64.20 \\
& SparseGPT & 37.20 & 12.80 & 25.00 \\
& Wanda & \underline{43.40} & \underline{26.22} & \underline{34.81} \\
\rowcolor{Gainsboro!40} & Ours & \textbf{44.00} & \textbf{28.05} & \textbf{36.02} \\
\midrule
\multirow{4}{*}{Qwen3-14B} & Base & 71.60 & 55.49 & 63.54 \\
& SparseGPT & 25.61 & \textbf{49.40} & 37.50 \\
& Wanda & \textbf{56.00} & 44.51 & \underline{50.26} \\
\rowcolor{Gainsboro!40} & Ours & \underline{55.00} & \underline{46.34} & \textbf{50.67} \\
\midrule
\multirow{4}{*}{DeepSeek-R1-Distill-Qwen-7B} & Base & 47.60 & 10.98 & 29.29 \\
& SparseGPT & 25.40 & 0.00 & 12.70 \\
& Wanda & \underline{27.40} & \underline{1.20} & \underline{14.30} \\
\rowcolor{Gainsboro!40} & Ours & \textbf{30.00} & \textbf{1.80} & \textbf{15.90} \\
\midrule
\multirow{4}{*}{DeepSeek-R1-Distill-Qwen-14B} & Base & 60.00 & 31.71 & 45.85 \\
& SparseGPT & 41.80 & 2.44 & 22.12 \\
& Wanda & \underline{43.60} & \underline{3.05} & \underline{23.32} \\
\rowcolor{Gainsboro!40} & Ours & \textbf{44.80} & \textbf{4.27} & \textbf{24.53} \\
\bottomrule
\end{tabular}
}
\vspace{-0.1in}
\end{table}

\begin{table}[!htbp]
\centering
\caption{Performance comparison of base and pruned vision-language models on multimodal benchmarks at 50\% unstructured sparsity. Values are shown as percentages. Bold indicates best performance among pruning methods, underlined indicates second best.}
\label{tab:vlm_comprehensive}
\resizebox{0.52\textwidth}{!}{%
\begin{tabular}{@{}ll*{4}{c}@{}}
\toprule
\textbf{Model} & \textbf{Method} & \rotatebox{0}{\textbf{MMMU}} & \rotatebox{0}{\textbf{MTVQA}} & \rotatebox{0}{\textbf{Mean}} \\
\midrule
\multirow{4}{*}{Qwen2.5-VL-3B} & Base & 48.60 & 24.34 & 36.47 \\
& SparseGPT & 30.33 & 19.66 & 25.00 \\
& Wanda & \textbf{37.50} & \underline{19.96} & \underline{28.73} \\
\rowcolor{Gainsboro!40} & Ours & \underline{36.81} & \textbf{20.97} & \textbf{28.89} \\
\midrule
\multirow{4}{*}{Qwen2.5-VL-7B} & Base & 49.55 & 29.59 & 39.57 \\
& SparseGPT & 39.33 & \textbf{27.14} & 33.24 \\
& Wanda & \underline{42.66} & 25.95 & \underline{34.31} \\
 \rowcolor{Gainsboro!40} & Ours & \textbf{43.77} & \underline{26.09} & \textbf{34.93} \\
\bottomrule
\end{tabular}%
}
\vspace{-0.1in}
\end{table}

\begin{table}[!htbp]
\centering
\caption{Ablation on the pruning metrics.}
\label{tab:gradient_only}
\resizebox{0.4\textwidth}{!}{%
\begin{tabular}{lccc}
\toprule
Method & \multicolumn{1}{c}{Sparsity} & \multicolumn{1}{c}{7B} & \multicolumn{1}{c}{13B} \\ \midrule 
Magnitude & 0.5 & \multicolumn{1}{c}{16.03} & 6.83 \\ 
SparseGPT & 0.5 & \multicolumn{1}{c}{7.00} & 6.03 \\ 
Wanda & 0.5 & \multicolumn{1}{c}{\bf 6.92} & 5.97 \\ 
$\left|\textbf W\right|\cdot\left\|\textbf G\right\|_1$ (Ours) & 0.5 & \multicolumn{1}{c}{7.17} & 6.15 \\ 
$\left|\textbf W\right|\cdot\left\|\textbf G\right\|_2$ (Ours) & 0.5 & \multicolumn{1}{c}{7.09} & \bf 5.96 \\
\bottomrule
\end{tabular}
}
\vspace{-0.1in}
\end{table}

\begin{table}[!htbp]
\centering
\caption{Pruning metric on weight, gradient, activation.}
\label{tab:prune_metric}
\resizebox{0.46\textwidth}{!}{%
\begin{tabular}{lcc}
\toprule
Method & \multicolumn{1}{l}{Sparsity} & \multicolumn{1}{l}{Perplexity}\\ \midrule 
$\left|\textbf W\right|\cdot \left| \textbf G_{acc}\right|$ & 0.5 & 119.72 \\
$\left|\textbf W\right| \cdot \left\|\textbf G\right\|_1$ & 0.5 & 7.17 \\
$\left|\textbf W\right|\cdot \left\|\textbf G\right\|_2$ & 0.5 & 7.09 \\
$(\left|\textbf W\right|\cdot \left\|\textbf X\right\|_2)^2$ + $\alpha \cdot \left|\textbf W\right| \cdot \left\|\textbf G\right\|_1$ & 0.5 & 6.90 \\
$(\left|\textbf W\right|\cdot \left\|\textbf X\right\|_2)^2$ + $\alpha \cdot \left|\textbf W\right|\cdot \left\|\textbf G\right\|_2$ & 0.5 & 6.88 \\
$(\left|\textbf W\right|\cdot \left\|\textbf X\right\|_2)^2$ - $\alpha \cdot \left|\textbf W\right|\cdot\left\|\textbf G\right\|_1$ & 0.5 & 9743.65 \\ 
$(\left|\textbf W\right|\cdot \left\|\textbf X\right\|_2)^2$ - $\alpha \cdot \left|\textbf W\right|\cdot\left\|\textbf G\right\|_2$ & 0.5 & 9377.00 \\
$\left|\textbf W\right|\cdot\left\|\textbf X\right\|_2$ + $\alpha \cdot \left|\textbf W\right|\cdot\left\|\textbf G\right\|_1$ & 0.5 & \textbf{6.86} \\
$\left|\textbf W\right|\cdot\left\|\textbf X\right\|_2$ + $\alpha \cdot \left|\textbf W\right|\cdot\left\|\textbf G\right\|_2$ & 0.5 & 6.89 \\
\bottomrule
\end{tabular}
}
\vspace{-0.15in}
\end{table}

\subsection{Zero-Shot Tasks}

In addition to our perplexity evaluations, we further assess the performance of our method across six Zero-shot common-sense tasks included in the Eleuther AI lm-evaluation-harness library~\citep{eval-harness}: BoolQ~\citep{Clark2019BoolQET}, RTE~\citep{Wang2018GLUEAM}, HellaSwag~\citep{Zellers2019HellaSwagCA}, WinoGrande~\citep{Sakaguchi2019WinoGrande}, ARC-easy~\citep{Clark2018ThinkYH}, and OBQA~\citep{Mihaylov2018CanAS}. 
As noted by earlier work~\citep{Dettmers2022TheCF, Frantar2023SparseGPTML}, zero-shot evaluation on these tasks is known to be noisy but aggregate performance across multiple tasks enhances interpretability. 

Our comprehensive results for these tasks are presented in Table {\ref{tab:harness_combined}}, where models are pruned to 50$\%$ unstructured sparsity. {We evaluate our method across multiple model families including LLaMA-1~\citep{touvron2023llama} (7B, 13B, 30B), LLaMA-2~\citep{touvron2023llama2} (13B, 70B), LLaMA-3.1~\citep{grattafiori2024llama} (8B), Qwen2.5~\citep{yang2025qwen3} (7B), and Qwen3~\citep{yang2025qwen3} (14B).} Notably, while our proposed \texttt{GBLM-Pruner} outperforms both Wanda and SparseGPT in terms of perplexity, a consistent trend is not observed across all the individual tasks,  which aligns with existing literature~\citep{Frantar2023SparseGPTML, Dettmers2022TheCF}. However, the mean accuracy across all six tasks surpasses the performance of both SparseGPT and Wanda for most of the models. This observation aligns with our findings from the perplexity evaluation, suggesting the robustness and effectiveness of our approach.

\vspace{-0.06in}
\subsection{K-Shot Reasoning}
\vspace{-0.06in}
We evaluate \texttt{GBLM-Pruner} on STEM/general benchmarks: MMLU~\citep{hendrycks2020measuring} (5-shot), GPQA~\citep{rein2024gpqa} (5-shot) and GPQA-D, MATH~\citep{hendrycks2021measuring} (4-shot), and GSM8K~\citep{cobbe2021training} (5-shot), at 50\% unstructured sparsity across LLaMA-3.1~\citep{grattafiori2024llama} (8B), Qwen3~\citep{yang2025qwen3} (8B, 14B), and Qwen2.5~\citep{yang2025qwen3} (7B). As shown in Table~\ref{tab:math_science_performance}, our method yields the highest mean accuracy on all models; e.g., on Qwen3-14B it reaches 47.58\% vs. Wanda 46.31\% and SparseGPT 45.82\%. Gains are most pronounced on MATH and GSM8K, indicating preserved multi-step STEM reasoning.

\vspace{-0.06in}
\subsection{Coding Tasks}
\vspace{-0.06in}
We assess code generation on MBPP~\citep{austin2021program} (3-shot) and HumanEval~\citep{chen2021evaluating} (0-shot) with 50\% sparsity, comparing against SparseGPT and Wanda on Qwen3~\citep{yang2025qwen3} (8B, 14B) and DeepSeek-R1-Distill-Qwen~\citep{guo2025deepseek} (7B, 14B). Table~\ref{tab:model_performance} shows \texttt{GBLM-Pruner} consistently achieves the best mean accuracy; for Qwen3-14B it obtains 50.67\% vs. Wanda 50.26\% and SparseGPT 37.50\%, with similar improvements on the DeepSeek-R1-Distill-Qwen models.

\vspace{-0.05in}
\subsection{Evaluating Vision–Language Models (VLMs)}\label{appendix-sec-vlm}
We extend to VLMs by pruning Qwen2.5-VL-3B and Qwen2.5-VL-7B to 50\% sparsity and evaluating on MMMU~\citep{yue2024mmmu} and MTVQA~\citep{tang2024mtvqa}. As summarized in Table~\ref{tab:vlm_comprehensive}, \texttt{GBLM-Pruner} attains the top mean scores on both models (28.89\% and 34.93\%), indicating that its gradient-based optimization effectively transfers to multimodal architectures.

\subsection{Ablation Study}\label{sec:abl}
\vspace{-0.06in}
\noindent{\bf Importance of Gradient}. To emphasize the role of gradient, we perform an ablation experiment as shown in Table \ref{tab:gradient_only}, wherein we only consider the Gradient-Weight term of the \texttt{GBLM-Pruner} pruning metric. Our experiments show a substantial enhancement over magnitude-based pruning when utilizing gradients solely with weights, evident in both LLaMA-2 7B and 13B models. Additionally, the performance of our metric closely aligns with that of Wanda and SparseGPT for LLaMA-2 13B model.

\noindent{\bf Pruning Metric}.
In Section \ref{sec:theory}, we revisited the OBS framework by incorporating the first order gradient which yields $\delta \bm{E}_m = \left(\bm{w}_{m}\left\|{x}_m\right\|_2\right)^{2}+\bm{w}_{m}\left(-\bm{g}_m\right)$ as the pruning metric. To start with, we experiment with different ways of estimating the gradient magnitude from the calibration samples. We evaluated three methods: gradient accumulation, $\ell_1$ norm and $\ell_2$ norm applied to the gradient across calibration samples. For this experiment, we only utilize the pruning metric based on gradient alone with weight for better interpretability. From our experiment, we observe that gradient accumulation yields the least favorable results as depicted in Table \ref{tab:prune_metric}. For a deeper understanding, we compared the pruning pattern of gradient accumulation with $\ell_1$ and $\ell_2$ norm which shows that gradient accumulation gives a noisy estimate of the gradient magnitude while $\ell_1$ and $\ell_2$ norm reveal more structured patterns. A comparison between gradient accumulation and $\ell_1$ norm-based aggregation is shown in Figure \ref{fig:l1_vs_accum_grad}. Based on this, we adopt $\ell_1$ and $\ell_2$ norm-based gradient estimation for subsequent analysis.

Subsequently, based on our theoretical pruning metric $\delta \bm{E}_m$, we experiment with two different ways of coupling the activations and gradients as shown in Table \ref{tab:prune_metric}. We observe that in the case of $(\left|\textbf W\right|\cdot \left\|\textbf X\right\|_2)^2 - \left|\textbf W\right|\cdot \left\|\textbf G\right\|_p$ the pruning metric is completely disrupted. While for $(\left|\textbf W\right|\cdot \left\|\textbf X\right\|_2)^2 + \left|\textbf W\right|\cdot \left\|\textbf G\right\|_p$ gradients and activations complement each other and brings out the best performance. But, upon closer examination, we observe that the square of the first activation term significantly outweighs the contribution of the second term involving gradients. Consequently, we remove the square factor from the first term and add a scaling factor denoted as $\alpha$ to the second gradient term, resulting in the formulation of our final pruning metric as $\left|\textbf W\right|\cdot \left\|\textbf X\right\|_2$ + $\alpha \cdot\left|\textbf W\right|\cdot \left\|\textbf G\right\|_p$. This pruning metric with $\ell_1$ norm-based gradient aggregation gives the best result for unstructured pruning across all models. We also conduct experiments to calibrate the scaling factor $\alpha$ as shown in Table \ref{tab:scale_factor}. We vary the scaling factor and examine how the LLaMA-2-7B pruned model perplexity changes. For a scaling factor is equal to 100, we get the best perplexity.

\noindent{\bf Sparsity Variation}. The objective of this ablation is to assess the robustness of our method across varying sparsity. For this, we compare the perplexity of the unstructured pruned model obtained by \texttt{GBLM-Pruner} to that of Wanda, SparseGPT, and magnitude pruning. We consider two distinct model sizes: a smaller LLaMA-2 13B model and a larger LLaMA-1 30B model, each is subjected to different degrees of sparsity. The results are shown in Figure~\ref{fig:sparsity_vs_ppl}. From the figure, it is evident that \texttt{GBLM-Pruner} exhibits a similar trend to SparseGPT and Wanda, showing a decline in performance as sparsity increases. However, \texttt{GBLM-Pruner} consistently outperforms all other baseline methods across various levels of sparsity for both models.

\noindent{\bf Dependence on Calibration Sample}.  \texttt{GBLM-Pruner} uses a set of calibration samples to calculate gradients and activations for the pruning metric. To understand the robustness of the pruned model to the calibration set, we conduct two ablations:

\textbf{(1)} Robustness to calibration set: We randomly sampled 5 different calibration sets with 128 samples each and prune LLaMA-2 7B model to 0.5 sparsity using \texttt{GBLM-Pruner}. The resultant pruned models have perplexities: 6.86, 6.87, 6.89, 6.86, and 6.87 respectively.

\textbf{(2)} Number of samples in the calibration set: here, we want to assess the influence of the calibration set size on the performance of \texttt{GBLM-Pruner}. For this, we prune the LLaMA-2 7B model using various calibration sets with the number of samples ranging from 1 to 512. The results are reported in Figure \ref{fig:sample_calb}. From the figure, we can observe that in contrast to SparseGPT, our method exhibits a relatively lower sensitivity to variations in the number of calibration samples.

\begin{figure}[!htbp]
\vspace{-0.1in}
\centering
\includegraphics[width=0.36\linewidth]{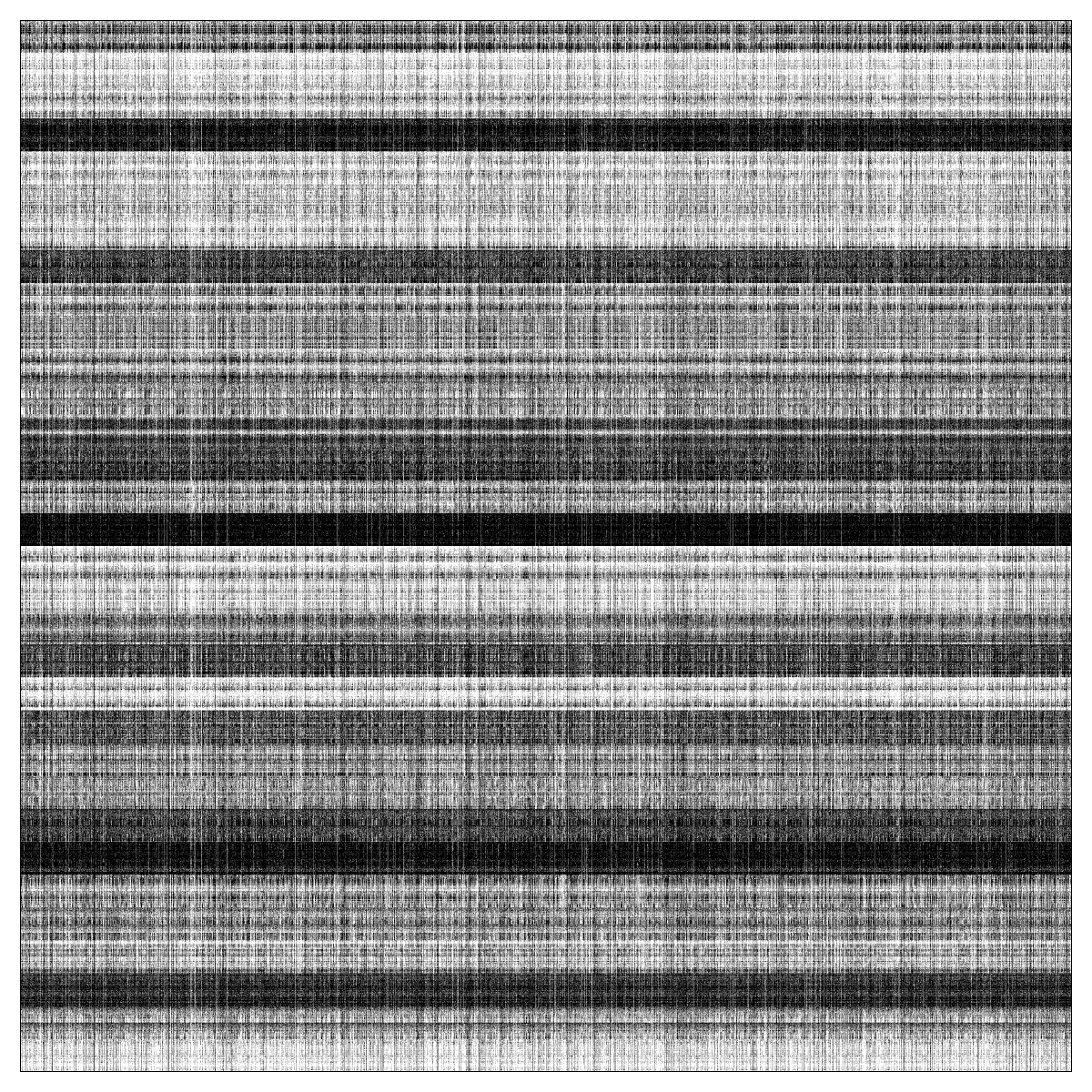}
\includegraphics[width=0.36\linewidth]{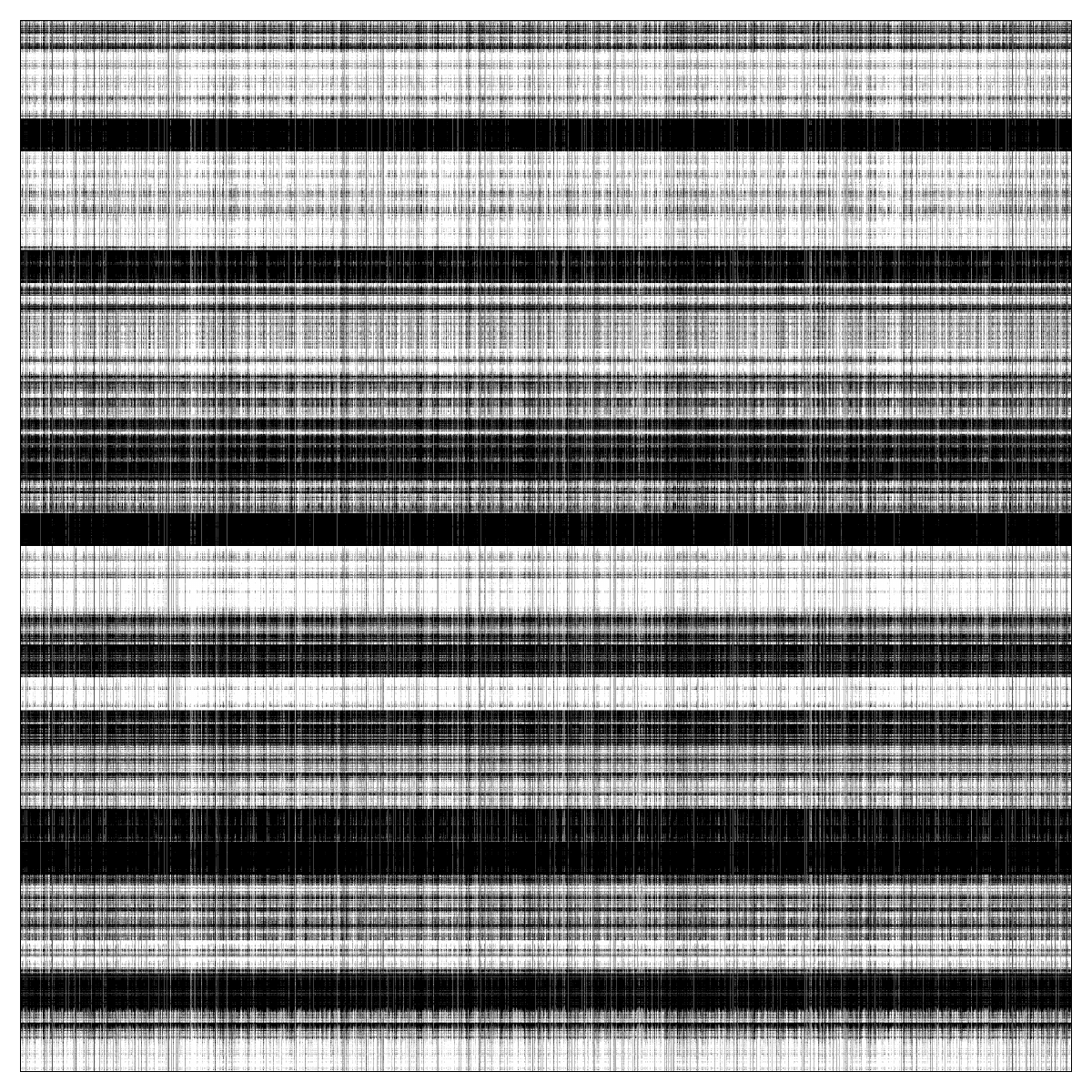}
  \vspace{-0.05in}
  \caption{Illustration of learned pruning pattern.}
  \label{fig:l1_vs_accum_grad}
\vspace{-0.15in}
\end{figure}

\subsection{Visualization of Pruned Pattern}
Figure~\ref{fig:l1_vs_accum_grad} visualizes 50\% pruning masks for the first layer’s key projection: left uses summation-aggregated gradients; right uses $\ell_1$-norm–aggregated gradients. In each subfigure, x denotes input dimensions and y denotes output dimensions. The sum-aggregated mask appears noisier, whereas the $\ell_1$-aggregated mask is sharper and more structured. After gradient aggregation, unstructured pruning reveals emergent structural patterns, reflecting geometric dependencies in LLM parameters aligned with gradient structure.

\subsection{Inference Time and Memory Comparison}
\texttt{GBLM-Pruner} has a one-time gradient pass and a layer-wise prune. Gradients require 71.3s/62GB (LLaMA1-7B) and 102.1s/96GB (LLaMA1-13B) on an A100 and are reusable across sparsity settings. The prune is $\mathcal{O}(d)$ like Wanda yet far faster than SparseGPT: 403.76ms vs.\ 256{,}732.95ms ($636\times$) on 7B and 561.30ms vs.\ 423{,}550.22ms ($754\times$) on 13B, with modest overhead vs.\ Wanda (257.04ms, 455.39ms). Speedups come from avoiding iterative weight updates/Hessian inverses. Also, 2:4 sparsity improves inference (overall $1.24\times$; per-layer $1.54$–$1.63\times$ on A6000) and reduces storage.

\begin{table}[!htbp]
\centering
\caption{Computation requirements on an A100 GPU.}
\resizebox{0.39\textwidth}{!}{%
\begin{tabular}{lcc}
\toprule
Model & Memory (GB) & Time (seconds) \\
\midrule
LLaMA-7B & 62 & 71.3 \\
LLaMA-13B & 96 & 102.1 \\
\bottomrule
\end{tabular}
}
\vspace{-0.1in}
\end{table}

\begin{table}[!htbp]
\centering
\caption{Pruning metric computation and weight removal time (milliseconds) on an A100 GPU.}
\resizebox{0.38\textwidth}{!}{%
\begin{tabular}{lcc}
\toprule
Method & LLaMA-7B & LLaMA-13B \\
\midrule
SparseGPT & 256,732.95 & 423,550.22 \\
Wanda & 257.04 & 455.39 \\
GBLM-Pruner & 403.76 & 561.30 \\
\bottomrule
\end{tabular}
}
\vspace{-0.1in}
\end{table}

\begin{table}[!htbp]
\centering
\caption{Inference speedup for 2:4 sparsity on an A6000 GPU across transformer layers of LLaMA-1.}
\resizebox{0.38\textwidth}{!}{%
\begin{tabular}{lccc}
\toprule
LLaMA-1 layer & Dense & 2:4 & Speedup \\
\midrule
q/k/v/o\_proj & 3.49 & 2.14 & 1.63$\times$ \\
up/gate\_proj & 9.82 & 6.10 & 1.61$\times$ \\
down\_proj & 9.92 & 6.45 & 1.54$\times$ \\
\bottomrule
\end{tabular}
}
\vspace{-0.1in}
\end{table}
}

\section{Related Work}

\vspace{-0.05in}

Large Language Models (LLMs) based on transformer architecture~\citep{Vaswani2017AttentionIA} have ushered in a transformative era in the realm of natural language processing, achieving outstanding success. Their consistent and remarkable performance spans a wide array of tasks~\citep{brown2020language, Chung2022ScalingIL, touvron2023llama, touvron2023llama2, Rozire2023CodeLO, openai2023gpt4, anil2023palm, grattafiori2024llama, yang2025qwen3, guo2025deepseek, team2026kimi, zeng2026glm, liu2026benchmarking}. For a long time, pruning has been identified as a powerful technique for reducing the size or complexity of a model by removing unnecessary or redundant components~\citep{lecun1989optimal,hassibi1993optimal}. 
Pruning can be divided into structured and unstructured pruning. Structured pruning targets at removing a set of weights from a network at once such as channels or layers to reduce the model size and complexity while maintaining the network structure intact. In the realm of pruning LLMs, several studies~\citep{Frantar2022OptimalBC, Frantar2023SparseGPTML, Sun2023ASA, lee2025unseen, yang2026lsa, yin2026duogpt, zou2026permllm, yu2026efficient} have been undertaken in this area. \textcolor{black}{Gradient-based importance estimation has proven effective in neural network pruning, as surveyed comprehensively by~\citet{hoefler2021sparsity}. Early approaches utilized activation gradients during training~\citep{ye2020accelerating} and position-based gradient scaling~\citep{kim2020position,lv2024fgp}. Recent gradient-based methods for LLMs focus on structured pruning, employing first-order gradients with Hessian approximations~\citep{ma2023llm}, forward-pass gradient estimation~\citep{cheng2024mini}, second-order curvature information~\citep{van2023llm,kurtic2022optimal,singh2020woodfisher}, regional gradients~\citep{yang2025wanda++}, gradient-guided layer pruning~\citep{huang2026gradpruner}, and LoRA-aware approaches~\citep{yu2025prunedlora,zhang2024loraprune}. While prior gradient-based methods focus on structured pruning or require task-specific finetuning or weight update, we introduce the first weight update-free gradient-based solution for unstructured and semi-structured LLM pruning that maintains zero-shot generalization capabilities.} Our work provides a unique angle from {\em gradient} along this direction\textcolor{black}{, combining first-order gradients with activation magnitudes in a computationally efficient manner}.

\section{Conclusion}

We presented a gradient-based pruning approach \texttt{GBLM-Pruner} for LLMs. Our approach performs in a training-free manner and applies gradient-based statistical magnitude to discern and selectively prune the model’s parameters, enabling substantial reductions in model size while preserving the model’s predictive accuracy. The proposed approach has surpassed all previous LLM pruning methods in terms of perplexity, zero-shot performance and interpretability, marking a pivotal advancement in the field. We also provided theoretical analyses on how gradients help identify the importance of weights in LLMs.

\section*{Ethics Statement}
The primary goal of this research is to improve the efficiency of LLMs by proposing a new network pruning approach. Any potential ethical issues of pretrained LLMs, such as harmful biases, privacy and spreading disinformation, are likely to still exist after pruning. 

\section*{Acknowledgments}

This work is supported by the MBZUAI-WIS Joint Program for Artificial Intelligence Research.

\bibliography{main}
\bibliographystyle{apalike}

\newpage 
\clearpage
\appendix

\section*{\Large{Appendix}}

\etocdepthtag.toc{mtappendix}
\etocsettagdepth{mtchapter}{none}
\etocsettagdepth{mtappendix}{subsection}
\tableofcontents
\newpage

\section{Vision Transformers}\label{appendix-sec-vit}

To assess the generalizability of our method across models with different input modalities, we conduct experiments on the ViT-B model. We compare the performance of the pruned model obtained using \texttt{GBLM-Pruner} with those obtained through magnitude pruning and the Wanda method.  We use 4,096 random samples from ImageNet-1k training set as our calibration data, and subsequently, we evaluate the pruned models on the standard ImageNet-1k classification task. The results of these evaluations are presented in Table \ref{tab:ViT-base_results}. We can see that our method outperforms both Wanda and magnitude pruning, particularly when dealing with higher levels of sparsity.

\begin{table}[!ht] 
\centering
\setlength{\tabcolsep}{4.0pt}
\caption{ViT-B model pruning.}
\vspace{1.ex}
\label{tab:ViT-base_results}
\begin{tabular}{ccccc}
\toprule
Sparsity & Wanda & Magnitude & Ours $\ell_1$ &  Ours $\ell_2$ \\ \midrule 
0 & 75.40 & 75.40 & 75.40 & 75.40 \\ 
0.5 & 64.54 & 59.48 & 64.64 & \bf 64.86 \\ 
0.6 & 43.65 & 29.98 & 44.15 & \bf 44.23 \\ 
0.7 & 7.92 & 1.88 & \bf 8.89 & 8.02 \\ 
0.8 & 0.20 & 0.18 & \bf 0.32 & 0.24 \\ 
\bottomrule
\end{tabular}
\vspace{-0.1in}
\end{table}

\section{\bf Difference to Previous Gradient-based Methods}
Although the use of gradients has been studied in the context of pruning, earlier methods ~\citep{molchanov2016pruning,sanh2020movement} used gradients in the context of transfer learning to obtain a pruned model that preserves the accuracy of the downstream task. This work is the first attempt to study the use of gradients for one-shot pruning of language models with billions of parameters while maintaining the zero-shot generalization capabilities of the language models to diverse downstream tasks. Additionally, our proposed method does not require weight update, which makes our proposed method computationally efficient and applicable for large language models with billions of parameters like LLaMA-1-30B and LLaMA-2-70B.

\section{Baselines}
We compare our proposed method against three pruning baselines:
\begin{itemize}
\addtolength{\itemsep}{-0.05in}
\item \textbf{Magnitude pruning}: Magnitude pruning~\citep{han2015learning} is a simple and scalable pruning method where the importance of LLM weights is decided based on the absolute value of their magnitude. Following \cite{gale2019state} and \cite{sanh2020movement}, we conduct a layer-wise comparison of model weights, subsequently removing those with smaller magnitudes.
\item \textbf{SparseGPT}: SparseGPT~\citep{Frantar2023SparseGPTML} is based on the second-order Optimal Brain Surgeon framework \citep{hassibi1993optimal}. It optimizes the accurate Optimal Brain Surgeon framework and introduces the first accurate one-shot pruning method that works efficiently at the scale of billions of parameters.
\item \textbf{Wanda}: Wanda~\citep{Sun2023ASA} proposed a simple pruning metric and showed the importance of activations in addition to weight magnitude while selecting weights for pruning. Unlike previous algorithms, it does not require any weight update of the remaining weights.
\end{itemize}

\section{Evaluation Metric}
Perplexity and Zero-shot Evaluation on Harness are two well-established metrics for evaluating compressed models:
\begin{itemize}
    \addtolength{\itemsep}{-0.05in}
    \item \textbf{Perplexity}: Following previous work on model compression both in case of quantization \citep{Dettmers2022TheCF, Yao2022ZeroQuantEA} and pruning \citep{Frantar2022OptimalBC, Sun2023ASA, Frantar2023SparseGPTML} we used perplexity as an evaluation metric to compare the pruned models. Perplexity is a stable, robust and challenging metric that is suited for evaluating the accuracy of compression methods. We used the WikiText~\citep{Merity2016PointerSM} validation set for computing perplexity.
    \item \textbf{Zero-Shot Evaluation on Harness Benchmarks}: To complement perplexity, we provided the evaluation of the pruned model on the publicly available Eleuther AI LM Harness benchmark~
    \citep{eval-harness} for additional interpretability. 
    We conducted evaluations on five standard common-sense reasoning tasks, including RTE \citep{Wang2018GLUEAM}, HellaSwag \citep{Zellers2019HellaSwagCA}, WinoGrande \citep{Sakaguchi2019WinoGrande}, ARC-easy \citep{Clark2018ThinkYH}, OBQA \citep{Mihaylov2018CanAS} and the BoolQ \citep{Clark2019BoolQET} reading comprehension task. Our evaluation primarily centers on assessing the pruned models' accuracy in comparison to the dense baseline, rather than emphasizing absolute numerical values.
\end{itemize}

\begin{table}[!htbp]
\centering
\vspace{-0.14in}
\caption{Ablation of scaling factor.}
\label{tab:scale_factor}
\begin{tabular}{lc}
\toprule
Scaling Factor, ($\alpha$) & \multicolumn{1}{l}{Perplexity}\\ \hline
0.001 & 6.920 \\
0.01 & 6.919 \\
0.1 & 6.921 \\
1 & 6.912 \\
10 & 6.890 \\
100 & \textbf{6.858} \\ 
1000 & 6.902 \\
10000 & 6.926 \\
100000 & 6.952 \\
\bottomrule
\end{tabular}
\end{table}

\section{Comparison Group}
Comparison group plays a pivotal role even in unstructured pruning. For \texttt{GBLM-Pruner}, we have experimented with 5 different comparison groups:
\begin{itemize}
    \item \texttt{Layer-wise}: With layer-wise pruning, weights within same layer are compared for pruning.
    \item \texttt{(input, 1)}:  For (input,1), weights connected within an input channel are grouped together for comparison.
    \item \texttt{(output, 1)}: Similarly in this approach, weights connected within an output channel are grouped together for comparison.
    \item \texttt{(input, 128)}:  This comparison group involves forming blocks of 128 input channels, and weights within each block are compared for pruning.
    \item \texttt{(output, 128)}: Similar to (input,128), here blocks of 128 channels are formed along the output dimension for pruning.
\end{itemize}

\section{LLaMA-Chat Models}
The LLaMA-2 series of models also includes fine-tuned chat versions. We sought to assess the generalization of our method to these chat models, specifically focusing on LLaMA-2-chat-7B and LLaMA-2-chat-13B as representative models. Similar to the pretrained LLaMA-2 series, our calibration data consisted of 128 samples, each comprising 2048 tokens from the C4 dataset. For evaluation purposes, we employed the Wiki-Text validation set.

Our approach to pruning was consistent with that applied to the pretrained LLaMA-2 models. We uniformly pruned every linear layer, except for the initial embedding layer and the final classification layer. We compare every weight of the linear layer on per output basis where pruning metric is compared within the output neuron. 

\begin{figure}[!htbp]
\centering
    \includegraphics[width=0.45\linewidth]{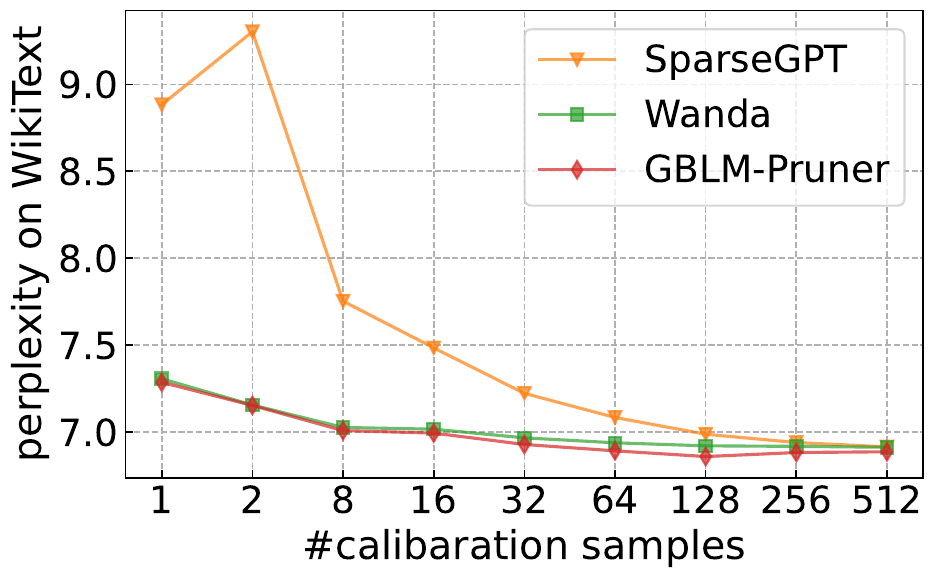}
    \vspace{-0.1in}
    \caption{Robustness to calibration samples.}
    \label{fig:sample_calb}
\end{figure}

\begin{figure}[!htbp]
\centering
  \begin{subfigure}{0.45\textwidth}  
    \centering
    \includegraphics[width=0.9\linewidth]{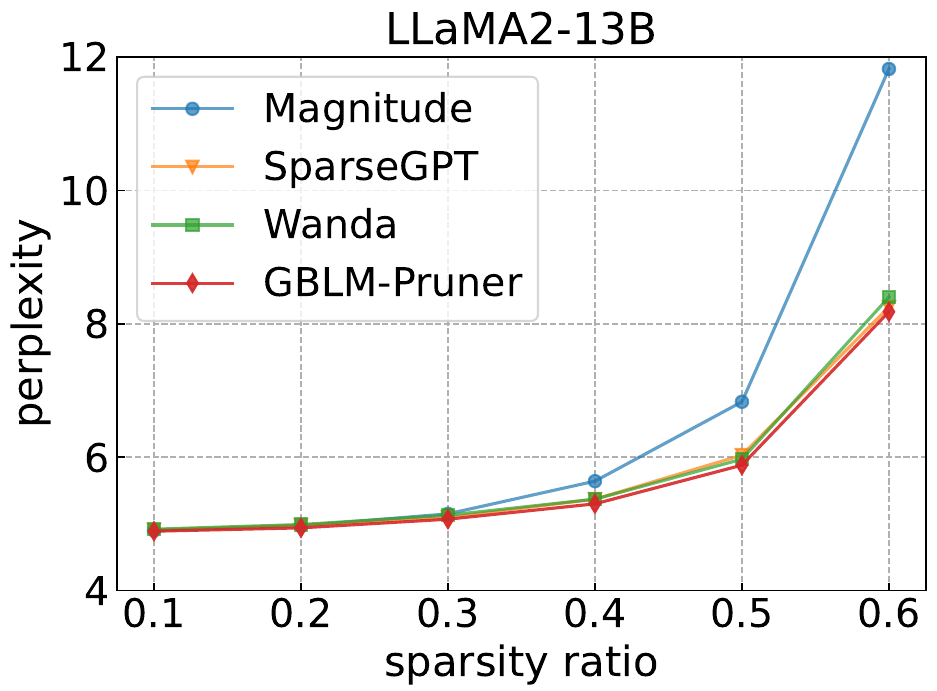} 
    \label{fig:subfig1}
  \end{subfigure}
  \begin{subfigure}{0.45\textwidth} 
    \centering
    \includegraphics[width=0.9\linewidth]{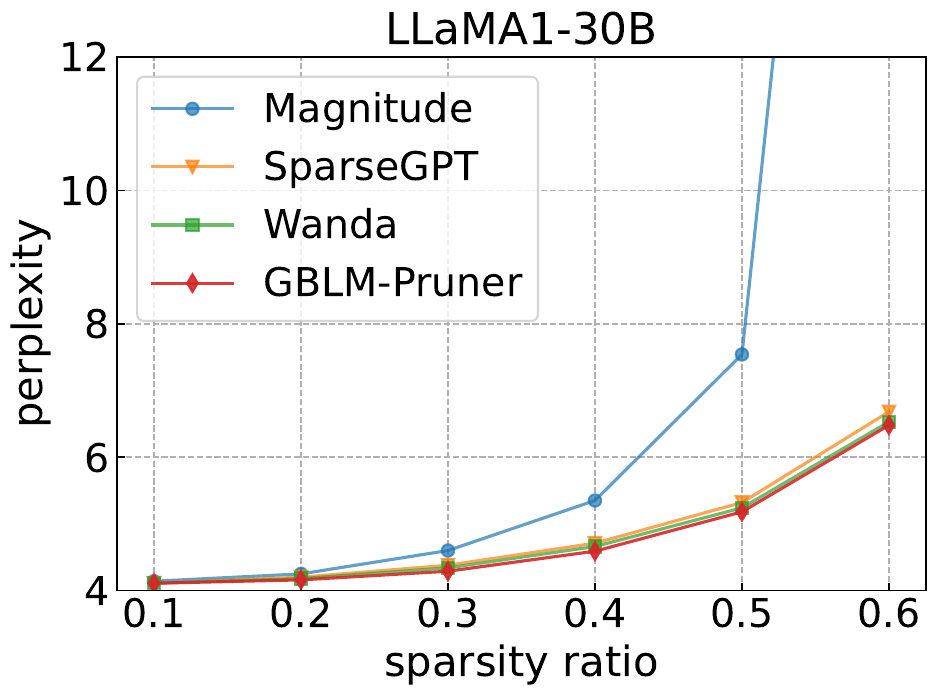} 
    \label{fig:subfig2}
  \end{subfigure}
    \vspace{-0.15in}
  \caption{Sparsity variation results for a large and a small model where we compare the performance of our method against other baseline methods.}
  \label{fig:sparsity_vs_ppl}
\end{figure}

The results are presented in Table \ref{tab:ppl_llama_chat}. Examining the table, we can discern that our method consistently delivers superior performance, particularly evident in unstructured pruning. When it comes to N$:$M sparsity pruning, although SparseGPT achieves the lowest perplexity, our pruning metric significantly outperforms Wanda by a substantial margin.
\begin{table}[!htbp]
\centering
\caption{WikiText validation perplexity of different pruning methods for LLaMA 2 chat models.}
\label{tab:ppl_llama_chat}
\resizebox{0.63\textwidth}{!}{
\begin{tabular}{lccc}
\toprule
Method & Sparsity & \multicolumn{1}{c}{LLaMA-2-7B-chat} & \multicolumn{1}{c}{LLaMA-13B-chat} \\ \hline
None & 0 & 7.08 & 6.11 \\
Magnitude & 0.5 & 22.82 & 8.49 \\ 
SparseGPT & 0.5 & 8.66 & 7.26 \\ 
Wanda & 0.5 & 8.78 & 7.50 \\ 
\texttt{GBLM-Pruner}$_{\ell2}$ & 0.5 & 8.52 & 7.27 \\ 
\texttt{GBLM-Pruner}$_{\ell1}$ & 0.5 & \textbf{8.40} & \textbf{7.10} \\ \hline
Magnitude & 2:4 & 45.95 & 11.14 \\ 
SparseGPT & 2:4 & \textbf{12.19} & \textbf{9.37} \\ 
Wanda & 2:4 & 14.45 & 10.25 \\ 
\texttt{GBLM-Pruner}$_{\ell2}$ & 2:4 & 13.74 & 9.85 \\ 
\texttt{GBLM-Pruner}$_{\ell1}$& 2:4 & 13.92 & 9.66 \\ \hline
Magnitude & 4:8 & 22.57 & 9.80 \\ 
SparseGPT & 4:8 & \textbf{10.02} & \textbf{8.01} \\ 
Wanda & 4:8 & 10.86 & 8.56 \\ 
\texttt{GBLM-Pruner}$_{\ell2}$ & 4:8 & 10.45 & 8.26 \\ 
\texttt{GBLM-Pruner}$_{\ell1}$ & 4:8 & 10.46 & 8.10 \\ \bottomrule
\end{tabular}
}
\end{table}

\section{OBS Weight Update}
In this study, our objective was to assess whether the OBS (Optimal Brain Surgeon) weight update method enhances the performance of our pruned model. We implemented the OBS weight update using the efficient approach proposed by SparseGPT~\citep{Frantar2023SparseGPTML}.

The results, presented in Table \ref{tab:obs}, indicate that the OBS weight update does not lead to an improvement in the performance of our pruned model

\begin{table}[!htbp]
\centering
\caption{OBS weight update.}
\label{tab:obs}
\resizebox{0.4\textwidth}{!}{
\begin{tabular}{lccc}
\toprule
\multicolumn{1}{l}{} &  & \multicolumn{2}{c}{Weight Update} \\ \hline
\multicolumn{1}{l}{Method} & \multicolumn{1}{c}{Datasplit} & \multicolumn{1}{c}{no} & \multicolumn{1}{c}{yes} \\ \hline
\multirow{2}{*}{Magnitude} & Calib & \multicolumn{1}{r}{18.14} & 12.93 \\  
 & Valid & \multicolumn{1}{r}{17.29} & 12.55 \\ \hline
\multirow{2}{*}{Wanda} & Calib & \multicolumn{1}{r}{7.52} & 7.61 \\
 & Valid & \multicolumn{1}{r}{7.26} & 7.36 \\ \hline
\multirow{2}{*}{Ours} & Calib & \multicolumn{1}{r}{7.54} & 7.64 \\ 
 & Valid & \multicolumn{1}{r}{7.26} & 7.39 \\ \bottomrule
\end{tabular}
}
\end{table}

\section{Correlations of Weights, Activations and Gradients.}  This section discusses an intuitive explanation of why gradient is essential. Weights are parameters in LLMs that are learned during the training process to minimize the loss function. They are fundamental in determining the strength of the connection between two neurons and subsequently the output of the network. Gradients of the loss with respect to weights, computed using an optimization algorithm like SGD~\citep{ruder2016overview}, are central to the learning process as they guide the updates made to the weights during training. On the otherhand, activations are the outputs of the neurons, typically computed as a weighted sum of inputs passed through an activation function. The activations are intrinsically impacted by the weights thus weight augmented with activation serves as a redundant indicator of weight importance. However, gradient being the guiding signal for the learning process serves as a valuable indicator by signalling the sensitivity of the loss to weight change and thus the importance of the weight in the pruning process.

\begin{table*}[t]
\centering
\caption{Pruning metric.}
\label{tab:prune_metric_app}
\resizebox{\textwidth}{!}{
\begin{tabular}{lcc|lcc}
\toprule
Method & \multicolumn{1}{c}{Sparsity} & \multicolumn{1}{c|}{Perplexity} &Method & \multicolumn{1}{c}{Sparsity} & \multicolumn{1}{c}{Perplexity} \\ \hline
$\left|\textbf W\right|\cdot \left|\textbf G_{acc}\right|$ & 0.5 & 119.72 & $(\left|\textbf W\right|\cdot \left\|\textbf X\right\|_2)^2$ + $\alpha \cdot \textbf W \cdot \textbf G_{acc}$ & 0.5 & 7.04\\ 
$\left|\textbf W\right| \cdot \left\|\textbf G\right\|_1$  & 0.5 & 7.17 & $(\left|\textbf W\right|\cdot \left\|\textbf X\right\|_2)^2$ + $\alpha \cdot \textbf W \cdot \left\|\textbf G\right\|_1$ & 0.5 & 180490.19\\ 
$\left|\textbf W\right|\cdot \left\|\textbf G\right\|_2$ & 0.5 & 7.09 & $(\left|\textbf W\right|\cdot \left\|\textbf X\right\|_2)^2$ + $\alpha \cdot \textbf W \cdot \left\|\textbf G\right\|_2$ & 0.5 & 91781.49 \\ 
$\left|\textbf W\right|\cdot \left\| \textbf X \right\|_2 \cdot \left|\textbf G_{acc}\right|$ & 0.5 & 69.59 & $(\left|\textbf W\right|\cdot \left\|\textbf X\right\|_2)^2$ - $\alpha \cdot \textbf W \cdot \textbf G_{acc}$ & 0.5 & 7.14 \\ 
$\left|\textbf W\right|\cdot \left\| \textbf X \right\|_2 \cdot \left\|\textbf G\right\|_1$  & 0.5 & 7.31 & $(\left|\textbf W\right|\cdot \left\|\textbf X\right\|_2)^2$ - $\alpha \cdot \textbf W \cdot \left\|\textbf G\right\|_1$ & 0.5 & 246846.28 \\ 
$\left|\textbf W\right|\cdot \left\| \textbf X \right\|_2\cdot \left\|\textbf G\right\|_2$  & 0.5 & 7.31 & $(\left|\textbf W\right|\cdot \left\|\textbf X\right\|_2)^2$ - $\alpha \cdot \textbf W \cdot \left\|\textbf G\right\|_2$ & 0.5 & 283620.75\\ 
$\left|\textbf W\right|\cdot \left\|\textbf X\right\|_2$ + $\alpha \cdot \left|\textbf W\right|\cdot\left|\textbf G_{acc}\right|$ & 0.5 & 6.92 & $(\left|\textbf W\right|\cdot \left\|\textbf X\right\|_2)^2$ + $\alpha \cdot \left|\textbf W\right|\cdot\left|\textbf G_{acc}\right|$ & 0.5 & 6.91\\ 
$\left|\textbf W\right|\cdot\left\|\textbf X\right\|_2$ + $\alpha \cdot \left|\textbf W\right|\cdot\left\|\textbf G\right\|_1$ & 0.5 & \textbf{6.86} & $(\left|\textbf W\right|\cdot \left\|\textbf X\right\|_2)^2$ + $\alpha \cdot \left|\textbf W\right| \cdot \left\|\textbf G\right\|_1$ & 0.5 & 6.90 \\ 
$\left|\textbf W\right|\cdot\left\|\textbf X\right\|_2$ + $\alpha \cdot \left|\textbf W\right|\cdot\left\|\textbf G\right\|_2$ & 0.5 & 6.89 & $(\left|\textbf W\right|\cdot \left\|\textbf X\right\|_2)^2$ + $\alpha \cdot \left|\textbf W\right|\cdot \left\|\textbf G\right\|_2$ & 0.5 & 6.88 \\ 
$\left|\textbf W\right|\cdot \left\|\textbf X\right\|_2$ - $\alpha \cdot \left|\textbf W\right|\cdot\left|\textbf G_{acc}\right|$ & 0.5 & 6.92 & $(\left|\textbf W\right|\cdot \left\|\textbf X\right\|_2)^2$ - $\alpha \cdot \left|\textbf W\right|\cdot\left|\textbf G_{acc}\right|$ & 0.5 & 6.94  \\ 
$\left|\textbf W\right|\cdot \left\|\textbf X\right\|_2$ - $\alpha \cdot \left|\textbf W\right|\cdot\left\|\textbf G\right\|_1$ & 0.5 & 1180.67 & $(\left|\textbf W\right|\cdot \left\|\textbf X\right\|_2)^2$ - $\alpha \cdot \left|\textbf W\right|\cdot\left\|\textbf G\right\|_1$ & 0.5 & 9743.65 \\ 
$\left|\textbf W\right|\cdot \left\|\textbf X\right\|_2$ - $\alpha \cdot \left|\textbf W\right|\cdot\left\|\textbf G\right\|_2$ & 0.5 & 7.10 & $(\left|\textbf W\right|\cdot \left\|\textbf X\right\|_2)^2$ - $\alpha \cdot \left|\textbf W\right|\cdot\left\|\textbf G\right\|_2$ & 0.5 & 9377.00 \\ \bottomrule
\end{tabular}
}
\end{table*}

\section{Optimal Brain Surgeon Considering gradient}
\label{app:obs}
As part of the theoretical justification for our proposed gradient-based metric, we revisited and redefined the OBS framework by incorporating considerations of the gradient information.  The complete derivation of this process is meticulously presented within this section.

The Taylor Series expansion of the error with respect to weight is:
\begin{equation}
\begin{aligned}
    \delta \bm{E} = \left(\frac{\partial \bm{E}}{\partial \bm w}\right)^{\top} \cdot \delta \bm w + \frac{1}{2} \delta \bm w^{\top}\cdot \textbf H  \cdot \delta \bm w + \mathcal{O} (\left| \left|\delta \bm w \right|\right|^{3})
\end{aligned} 
\label{eq:loss_taylor_exp_app}
\end{equation}
where $\bm E$ is the error or loss function and $\bm w$ is the weight vector for the neural network.  The symbol $\textbf H=\frac{\partial^2 \bm E}{\partial \bm w^2}$ denotes the Hessian Matrix, and the superscript $\top$ signifies vector transpose. Based on this we formulate the optimization problem for network pruning using both the first and second-order terms as depicted in Equation \ref{eq:prun_opt_obj_app}. Here, $\bm w_m$ is the pruned weight, $\delta \bm w$ is the change in weight magnitude for $\bm w_m$ and $\mathbf{e}_{m}$ is the unit vector in weight space corresponding to weight $\bm w_m$.

\begin{equation}
\begin{aligned}
    \min_{m}\Bigg\{\min_{\delta \bm w}\bigg(\left(\frac{\partial \bm{E}}{\partial \bm w}\right)^{\top}\!\cdot\delta \bm{w} \\
    + \frac{1}{2} \delta \bm{w}^{\top} \cdot \textbf{H} \cdot \delta \bm{w} \bigg) &\Big| \mathbf{e}_m^{\top} \cdot \delta \bm{w} + {w}_m = 0\Bigg\}
\end{aligned}  
\label{eq:prun_opt_obj_app}
\end{equation}

The Lagrangian formulation of the optimization problem is:

\begin{equation}
\begin{aligned}
    {\mathcal{L}} = \bm{g}^{\top} \cdot \delta \bm{w} + \frac{1}{2} \delta \bm{w}^{\top} \cdot \textbf{H} \cdot \delta \bm{w}  + \lambda\left( \mathbf{e}_m^{\top} \cdot \delta \bm{w} + {w}_m \right)
\end{aligned}  
\label{eq:lag_app}
\end{equation}

Now, differentiating Equation \ref{eq:lag_app} w.r.t $\lambda$

\begin{equation}
\begin{aligned}
\mathbf{e}_m^{\top} \cdot \delta \bm{w} + {w}_m = 0
\end{aligned}
\label{eq:diff_lamda}
\end{equation}

Differentiating w.r.t $\delta \bm w$

\begin{equation}
\begin{aligned}
& \bm{g}+\textbf{H} \cdot \delta \bm{w}+\lambda \mathbf{e}_{m}=0 \\
 \Rightarrow& \delta \bm{w}=-\textbf{H} ^{-1}\cdot (\lambda \mathbf{e}_{m} +\bm{g})
\end{aligned}
\label{eq:diff_delta_w}
\end{equation}

From \ref{eq:diff_lamda} and \ref{eq:diff_delta_w}, we have

\begin{equation}
\begin{aligned}
& \mathbf{e}_m^{\top}\left(-\textbf{H}^{-1}\cdot(\lambda \mathbf{e}_{m}+\bm{g})\right)+{w}_{m}=0 \\
& \Rightarrow -\lambda\left(\bm{H}^{-1}\right)_{m m}-\mathbf{e}_m^{\top} \cdot \textbf{H}^{-1} \cdot \bm{g}+{w}_m=0 \\
& \Rightarrow \lambda=\frac{{w}_m-\mathbf{e}_m^{\top} \cdot \textbf{H}^{-1} \cdot \bm{g}}{\left(\textbf{H}^{-1}\right)_{m m}} \\
\end{aligned}
\label{eq:lamda_cal}
\end{equation}

From \ref{eq:diff_delta_w} and \ref{eq:lamda_cal}, we get the optimal weight change $\delta \bm{w}$ as:
\begin{equation}
\begin{aligned}
&\delta \bm{w}=-\textbf{H}^{-1} \cdot \left(\frac{{w}_m-\mathbf{e}_m^{\top} \cdot \textbf{H}^{-1} \cdot \bm{g}}{\left(\textbf{H}^{-1}\right)_{m m}} \cdot \mathbf{e}_{m}+\bm{g}\right) \\
& =-\frac{{w}_{m}}{\left(\textbf{H}^{-1}\right)_{m m}} \textbf{H}^{-1} \cdot \mathbf{e}_{m}+\frac{\mathbf{e}_m^{\top} \cdot \textbf{H}^{-1} \cdot {g}}{\left(\textbf{H}^{-1}\right)_{m m}} \textbf{H}^{-1} \cdot \mathbf{e}_{m} -\textbf{H}^{-1} \cdot \bm{g} \\
\end{aligned}
\end{equation}

The increase in error on changing weight ${w}_m$ by $\delta \bm{w}$ is:
\begin{equation}
\begin{aligned}
    \delta \bm{E}_m = \bm{g}^{\top} \cdot \delta \bm{w} + \frac{1}{2} \delta \bm{w}^{\top} \cdot \textbf{H} \cdot \delta \bm{w}
\end{aligned}
\label{eq:saliency}
\end{equation}

Substituting the optimal value of $\delta \bm{w}$ in Equation \ref{eq:saliency} gives:
\begin{equation}
\begin{aligned} 
\delta \bm E_m &= \frac{w_{m}^{2}}{2\left(\textbf{H}^{-1}\right)_{mm}} - \frac{w_{m}\left(\bm{g}^{\top} \cdot \textbf{H}^{-1} \cdot \mathbf{e}_{m}\right)}{\left(\textbf{H}^{-1}\right)_{mm}} \\
&\quad + \frac{\left(\mathbf{e}_{m}^{\top} \cdot \textbf{H}^{-1} \cdot \bm{g}\right)^{2}}{2\left(\textbf{H}^{-1}\right)_{mm}} - \frac{1}{2} \bm{g}^{\top} \cdot \textbf{H}^{-1} \cdot \bm{g}
\end{aligned}
\label{eq:saliency_sol}
\end{equation}

\section{Different Pruning Metric}
In the ablation Section \ref{sec:abl}, we present an analysis of our pruning metric. Table \ref{tab:prune_metric_app} enumerates all the pruning metrics we explored and serves as a comprehensive consolidation of our study.

{\color{black}

\section{Higher Sparsity Evaluation}
\label{app:high_sparsity}

To evaluate the robustness of \texttt{GBLM-Pruner} at extreme sparsity levels, we conduct experiments at 0.7, 0.8, 0.9, and 0.95 sparsity ratios. Table~\ref{tab:high_sparsity_ppl} reports WikiText perplexity for Qwen3-14B and LLaMA-3.1-8B models. At all sparsity levels, we can observe that gradient information is extremely valuable: \texttt{GBLM-Pruner} achieves 24\% lower perplexity than Wanda on Qwen3-14B at 0.8 sparsity, and 52\% lower at 0.9 sparsity. Similar trends hold for LLaMA-3.1-8B, where our method achieves 43\% and 41\% lower perplexity at 0.8 and 0.9 sparsity, respectively. At extreme sparsity (0.95), models exhibit degraded performance, but \texttt{GBLM-Pruner} maintains a substantial advantage (30\% lower perplexity). These results demonstrate that gradient-based pruning becomes particularly important at higher sparsity levels where pruning decisions are most challenging.

\begin{table}[h]
\centering
\color{black}
\caption{WikiText perplexity at higher sparsity levels. Bold indicates best performance.}
\label{tab:high_sparsity_ppl}
\resizebox{0.5\columnwidth}{!}{
\begin{tabular}{lccc}
\toprule
\textbf{Model} & \textbf{Sparsity} & \textbf{Wanda} & \textbf{GBLM-Pruner} \\
\midrule
\multirow{4}{*}{Qwen3-14B} 
& 0.7 & 33.12 & \textbf{32.13} \\
& 0.8 & 254.90 & \textbf{192.85} \\
& 0.9 & 10902.73 & \textbf{5225.87} \\
& 0.95 & 16872.77 & \textbf{11845.89} \\
\midrule
\multirow{3}{*}{LLaMA-3.1-8B} 
& 0.7 & 113.59 & \textbf{111.48} \\
& 0.8 & 1480.43 & \textbf{835.78} \\
& 0.9 & 13155.59 & \textbf{7726.91} \\
\bottomrule
\end{tabular}
}
\end{table}

\section{Detailed Hessian Derivation}
\label{app:hessian_derivation}

This appendix provides the complete derivation connecting our Hessian approximation $\mathbf{H} \approx \mathbf{X}^{\top}\mathbf{X}$ to the original Optimal Brain Surgeon framework. We show how layer-wise analysis combined with diagonal approximations yields the form used in our pruning metric.

Following Hassibi and Stork~\citep{hassibi1993optimal}, for the squared error loss function under the $(t-o) \to 0$ approximation for well-trained networks, the Hessian of the global loss function can be expressed as (OBS Equation 11):
\begin{equation}
\mathbf{H} \approx \frac{1}{N}\sum_{k=1}^{N} \mathbf{J}^{[k]\top}\mathbf{J}^{[k]}
\label{eq:app_hessian_obs}
\end{equation}
where $\mathbf{J}^{[k]} = \frac{\partial \mathbf{F}(\mathbf{w}, \mathbf{x}^{[k]})}{\partial \mathbf{w}}$ is the Jacobian of network outputs with respect to weights for sample $k$, and $N$ is the number of calibration samples.

Consider a single linear layer within the network with transformation $\mathbf{a}^{[k]} = \mathbf{W}\mathbf{x}^{[k]}$, where the $i$-th output is $a^{[k]}_i = \sum_j W[i,j] \cdot x^{[k]}_j$. Applying the chain rule, the derivative of network output $F_\ell$ with respect to weight $W[i,j]$ is:
\begin{equation}
\frac{\partial F_\ell(\mathbf{w}, \mathbf{x}^{[k]})}{\partial W[i,j]} = \frac{\partial F_\ell}{\partial a^{[k]}_i} \cdot \frac{\partial a^{[k]}_i}{\partial W[i,j]} = B^{[k]}_{\ell i} \cdot x^{[k]}_j
\label{eq:app_chain_rule}
\end{equation}
where $B^{[k]}_{\ell i} := \frac{\partial F_\ell(\mathbf{w}, \mathbf{x}^{[k]})}{\partial a^{[k]}_i}$ represents the backpropagated sensitivity of output $\ell$ to the pre-activation $a_i$ of our layer.

Substituting Equation~\ref{eq:app_chain_rule} into the Hessian formula from Equation~\ref{eq:app_hessian_obs}, the Hessian element corresponding to weights $W[i,j]$ and $W[i',j']$ becomes:
\begin{equation}
\begin{aligned}
H_{(i,j),(i',j')} &\approx \frac{1}{N}\sum_{k=1}^{N} \sum_{\ell} \left(B^{[k]}_{\ell i} x^{[k]}_j\right)\left(B^{[k]}_{\ell i'} x^{[k]}_{j'}\right) \\
&= \frac{1}{N}\sum_{k=1}^{N} x^{[k]}_j x^{[k]}_{j'} \underbrace{\sum_{\ell} B^{[k]}_{\ell i} B^{[k]}_{\ell i'}}_{M^{[k]}_{ii'}}
\end{aligned}
\label{eq:app_hessian_full}
\end{equation}
where $M^{[k]}_{ii'} = \sum_{\ell} B^{[k]}_{\ell i} B^{[k]}_{\ell i'}$ captures all backpropagation information and equals $\frac{\partial^2 E}{\partial a^{[k]}_i \partial a^{[k]}_{i'}}$, the second derivative of loss with respect to pre-activations.

We now introduce two simplifying assumptions to make the Hessian computation tractable. First, following LeCun et al.~\citep{lecun1989optimal} (Optimal Brain Damage), we assume that interactions between different output neurons in the same layer are negligible:
\begin{equation}
M^{[k]}_{ii'} \approx 0 \quad \text{for } i \neq i'
\label{eq:app_assumption_a1}
\end{equation}
This diagonal approximation across neurons gives a block-diagonal Hessian structure in neuron index $i$, substantially reducing computational complexity.

Second, we assume that the curvature terms $M^{[k]}_{ii} = \frac{\partial^2 E}{\partial a^{[k]}_i{}^2}$ are approximately constant across neurons and samples within a layer:
\begin{equation}
M^{[k]}_{ii} \approx c \quad \text{for all } i, k
\label{eq:app_assumption_a2}
\end{equation}
where $c$ is a constant. This is a strong assumption, as different neurons may have different curvature depending on their role in the network. However, computing per-neuron curvature would require expensive second-order computations. Moreover, this approximation is implicitly made in prior work such as Wanda~\citep{Sun2023ASA}, which uses $|\mathbf{W}[i,j]| \cdot \|\mathbf{X}[:,j]\|_2$ without any curvature terms.

Under these two assumptions, Equation~\ref{eq:app_hessian_full} simplifies to:
\begin{equation}
H_{(i,j),(i',j')} \approx c \cdot \delta_{ii'} \left[\frac{1}{N}\sum_{k=1}^{N} x^{[k]}_j x^{[k]}_{j'}\right] = c \cdot \delta_{ii'} \cdot [\mathbf{X}^\top\mathbf{X}]_{jj'}
\label{eq:app_hessian_simplified}
\end{equation}
where $\delta_{ii'}$ is the Kronecker delta.

Further applying the diagonal approximation in the input dimension $j$ (standard from Optimal Brain Damage), we consider only diagonal elements:
\begin{equation}
H_{(i,j),(i,j)} \approx c \cdot \sum_{k=1}^{N} (x^{[k]}_j)^2 = c \cdot \|\mathbf{X}[:,j]\|_2^2
\label{eq:app_hessian_diagonal_final}
\end{equation}

In compact notation for the entire layer, this can be written as:
\begin{equation}
\mathbf{H} \approx c \cdot \operatorname{diag}(\|\mathbf{X}[:,j]\|_2^2, \; 1 \leq j \leq d_{\text{in}})
\label{eq:app_hessian_compact}
\end{equation}

The constant $c$ depends on the loss function and layer properties, but for weight ranking purposes, it can be absorbed into the overall scaling. The inverse Hessian diagonal element is proportional to $[\mathbf{H}^{-1}]_{mm} \propto \frac{1}{\|\mathbf{x}_m\|_2^2}$, which directly leads to the first term in our pruning metric derivation (Equation~\ref{eq:lag_soln_first_term} in the main text). The empirically determined scaling factor $\alpha$ in our final metric (Equation~\ref{eq:practical_metric}) accounts for both this constant and the relative magnitudes of gradient and activation contributions. Our extensive empirical results across 15+ model families (Tables 1-6 in the main text) demonstrate that the resulting pruning metric performs effectively in practice, suggesting that the diagonal and uniform curvature approximations provide sufficient signal for identifying important weights.

}

\end{document}